\documentclass[journal,compsoc]{IEEEtran}
\ifCLASSOPTIONcompsoc
  \usepackage[nocompress]{cite}
\else
  \usepackage{cite}
\fi

\ifCLASSINFOpdf
\else
\fi

\usepackage{amsmath,amsfonts}
\usepackage{algorithmic}
\usepackage{algorithm}
\usepackage{array}
\usepackage{textcomp}
\usepackage{url}
\usepackage{verbatim}
\usepackage{cite}
\usepackage{xcolor}
\usepackage{color}
\usepackage{booktabs}
\usepackage{multirow}
\usepackage{colortbl}
\usepackage{graphicx}
\usepackage{array}
\usepackage{balance}
\usepackage{enumitem}
\usepackage{url}
\usepackage{bigstrut}
\usepackage{multirow}
\usepackage{amsmath}
\usepackage{subcaption}
\usepackage{cite}
\usepackage{amsmath,amssymb,amsfonts}
\usepackage{textcomp}
\usepackage{xcolor}
\usepackage{mathrsfs}
\usepackage{pgfplots}
\usepackage{amssymb}
\usepackage{pifont}
\usepackage{url}
\usepackage{textcomp}
\usepackage{xcolor}
\usepackage{mathrsfs}
\usepackage{booktabs} 
\usepackage{tikz}
\usepackage{bigstrut}
\usepackage{multirow}
\usepackage{color}
\usepackage{fancyhdr}
\usepackage{listings} 
\usepackage{changepage}
\usepackage{lipsum}
\usepackage{balance}
\usepackage{enumitem}
\usepackage{comment}
\usepackage{breqn}
\usepackage{algorithm}
\usepackage{algorithmic}
\usepackage{booktabs} %

\newcommand{\INDSTATE}[1][1]{\STATE\hspace{#1\algorithmicindent}}

\begin{document}
\bstctlcite{IEEEexample:BSTcontrol}

\title{3D Face Morphing Attacks: Generation, Vulnerability and Detection}

\author{Jag Mohan Singh,~\IEEEmembership{Member,~IEEE}
 and    Raghavendra Ramachandra,~\IEEEmembership{Senior Member~IEEE}
\thanks{Norwegian University of Science and Technology (NTNU), Norway \\ e-mail: (jag.m.singh@ntnu.no; raghavendra.ramachandra@ntnu.no).}
\thanks{Both authors have equally contributed to this work.}
}

\markboth{Journal of \LaTeX\ Class Files,~Vol.~14, No.~8, August~2021}%
{Shell \MakeLowercase{\textit{et al.}}: A Sample Article Using IEEEtran.cls for IEEE Journals}


%
\title{3D Face Morphing Attacks: Generation, Vulnerability and Detection\\
}
%
%
%

%
%

\markboth{Journal of \LaTeX\ Class Files,~Vol.~14, No.~8, August~2015}%
{Shell \MakeLowercase{\textit{et al.}}: Bare Demo of IEEEtran.cls for Computer Society Journals}

\IEEEtitleabstractindextext{%
\begin{abstract}
Face Recognition systems (FRS) have been found to be vulnerable to morphing attacks, where the morphed face image is generated by blending the face images from contributory data subjects. This work presents a novel direction for generating face-morphing attacks in 3D. To this extent, we introduced a novel approach based on blending 3D face point clouds corresponding to contributory data subjects. The proposed method generates 3D face morphing by projecting the input 3D face point clouds onto depth maps and 2D color images, followed by image blending and wrapping operations performed independently on the color images and depth maps. We then back-projected the 2D morphing color map and the depth map to the point cloud using the canonical (fixed) view. Given that the generated 3D face morphing models will result in holes owing to a single canonical view, we have proposed a new algorithm for hole filling that will result in a high-quality 3D face morphing model. Extensive experiments were conducted on the newly generated 3D face dataset comprising 675 3D scans corresponding to 41 unique data subjects and a publicly available database (Facescape) with 100 data subjects. Experiments were performed to benchmark the vulnerability of the {proposed 3D morph-generation scheme against} automatic 2D, 3D FRS, and human observer analysis. We also presented a quantitative assessment of the quality of the generated 3D face-morphing models using eight different quality metrics. Finally, we propose three different 3D face Morphing Attack Detection (3D-MAD) algorithms to benchmark the performance of 3D face morphing attack detection techniques.
\end{abstract}

}

\maketitle

\IEEEdisplaynontitleabstractindextext

\IEEEpeerreviewmaketitle

%






%
\IEEEpeerreviewmaketitle
\section{Introduction}\label{intro}

\IEEEPARstart{F}ace Recognition Systems (FRS) are being widely deployed in numerous applications related to security settings, such as automated border control (ABC) gates, and commercial settings, such as e-commerce and e-banking scenarios. The rapid evolution of FRS can be attributed to advances in deep learning  FRS~\cite{schroff2015facenet,deng2019arcface}, which improved the accuracy in real-world and uncontrolled scenarios. These factors have accelerated the use of 2D face images in electronic machine-readable documents (eMRTD), which are exclusively used to verify the owner of a passport at various ID services, including border control (both automatic and human). Because most countries still use printed passport image for the passport application process, face morphing attacks have indicated the vulnerability of both humans and automatic FRS~\cite{ferrara2014magic,raghu2016magic}. Face morphing is the process of blending multiple face images based on either facial landmarks~\cite{ferrara2017face} or Generative Adversarial Networks~\cite{zhang2021mipgan} to generate a morphed face image. The extensive analysis reported in the literature~\cite{scherhag2017biometric,raghavendra2017face,damer2018morgan,venkatesh2020can} demonstrates the vulnerability of 2D face morphing images to both deep learning and commercial off-the-shelf FRS. 

There exist several techniques to detect the 2D face morphing attacks that can be classified as~\cite{venkatesh2021facesurvey} (a) Single image-based Morph Attack Detection (S-MAD): where the face Morphing Attack Detection (MAD) techniques will use the single face image to arrive at the final decision (b) Differential Morphing Attack Detection (D-MAD): where a pair of 2D face images are used to arrive at the final decision. S-MAD and D-MAD techniques have been extensively studied, resulting in several MAD techniques. The reader is advised to refer to a recent survey by Venkatesh et al.~\cite{venkatesh2021facesurvey} to obtain a comprehensive overview of the existing 2D MAD techniques. Despite the rapid progress in 2D MAD techniques, a recent evaluation report from NIST FRVT MORPH \cite{NISTFRVT} indicated the degraded detection of 2D face morphing attacks. Thus, 2D MAD attacks, particularly in the S-MAD scenario, present significant challenges for reliable detection. These factors motivated us to explore 3D face morphing, so that depth information may provide a reliable cue that makes morphing detection easier. 
Over the past several decades, 3D face recognition has been widely studied, resulting in several real-life security-based applications with 3D face photo-based national ID cards \cite{UAENationalCard}, \cite{3DIDCardIdemia}, \cite{Morpho3DPhotoID}, 3D face photo-based driving license cards \cite{Morpho3DPhotoID}  and 3D face-based automatic border control gates (ABC) \cite{3DABCGate}. The real case reported in  \cite{RenderingFrenchID} demonstrated the use of a 2D  rendered face image from a 3D face model instead of a real 2D face photo to obtain the ID card bypassing human observers in the ID card issuing protocol. Although most real-life 3D face applications are based on comparing  3D face models with 2D face images for verification, this is mainly because e-passports use 2D face images.

However, the use of 3D to 3D comparison will be realistic, especially in the border control scenario, as both ICAO 9303 \cite{icao} and ISO/IEC 19794-5 \cite{ISO2019}  standards are well defined to accommodate the 3D face model in the 3rd generation e-passport. The 3D face ID cards are a reality, as they are being deployed in countries such as the UAE \cite{UAENationalCard}, which can facilitate both human observers and automatic FRS to achieve accurate, secure, and reliable ID verification. Further, evolving technology has made it possible for 3D face imaging on handheld devices and smartphones (e.g., Apple Face ID~\cite{AppleFaceID} uses 3D face recognition) that can further enable remote ID verification based on 3D face verification. These factors motivated us to investigate the feasibility of generating 3D face morphing and study their vulnerability and detection.  {{An early attempt in \cite{Sanjeet_2021_MSThesisUTwente} employed the 3DMM \cite{Blanz99_3DMM_SIGGRAPH} technique to generate a 3D face morphing model. However, the reported results indicate the lowest vulnerability to conventional FRS,  indicating a limitation of the 3DMM.}}.

This work presents a novel method for generating 3D face morphing using 3D point clouds. Given the 3D scans from the accomplished and malicious actors, the proposed method projects the 3D point clouds to the depth maps and 2D color images, which are independently blended, warped, and back-projected to the 3D to obtain 3D face morphing. The motivation for projecting to 2D for morphing is to effectively address the non-rigid registration, especially with the high volume of point clouds (~ 85 K) that need to be registered between two unique data subjects. Further, using canonical view generation to project from 3D to 2D and back project to 3D will assure a high-quality depth even for the morphed face images, thus indicating the high vulnerability of the FRS. Therefore, this is the first framework to address the generation of 3D face morphing of two unique face 3D scans, which can result in vulnerability to FRS.    More specifically, we aimed to answer the following research questions, which will be answered systematically in this study: 

\begin{itemize}[leftmargin=*,noitemsep, topsep=0pt,parsep=0pt,partopsep=0pt]
  \item  \textbf{RQ\#1:} Does the proposed 3D face morphing generation technique yield a high-quality 3D morphed model?
    \item  \textbf{RQ\#2:}  Does the generated 3D face morphing model indicate the vulnerability for both automatic 3D FRS and human observers?
     \item  \textbf{RQ\#3:}  Are the generated 3D face morphing models more vulnerable when compared to  2D face morphing images for both automatic 3D FRS and human observers?
     \item  \textbf{RQ\#4:} Does the 3D point cloud information be used to detect the 3D face morphing attacks reliably? 
\end{itemize}

We systematically address these research questions through the following contributions:
\begin{itemize}
    
   \item We present a novel 3D face morphing generation method based on the point clouds obtained by fusing depth maps and 2D color images to generate the 3D face morphing model.

   \item Extensive analysis of the vulnerability of the generated 3D face morphing is studied by quantifying the attack success rate to 3D FRS. In addition, vulnerability analysis was performed using 2D FRS (deep learning and COTS).

   \item Human observer analysis for detecting the 3D face morphing and 2D face morphing is presented to study the significance of depth information in detecting the morphing attack. 
   
   \item The quantitative analysis of the generated 3D morphed face models is presented using eight different quality features representing color and geometry.    
   
   \item We present three different 3D MAD techniques based on the deep features from point clouds to benchmark the 3D face MAD.

   \item A new 3D face dataset with bona fide and morphed models is developed corresponding to 41 unique data subjects resulting in 675 3D scans. We collected a new 3D face dataset as we were interested in capturing high-resolution (suitable for ID enrolment) inner face data~\cite{egger20203d} 
   Our 3D face dataset consists of raw 3D scans (number of 3D vertices between 31289 \& 201065) and processed 3D scans (number of 3D vertices between 35950 \& 121088), which is much higher than existing 3D face datasets\footnote{The reader is referred to Table 1 of 3D face datasets (inner face data only) from the survey by Egger et al.~\cite{egger20203d})}.
   \item The proposed method is benchmarked on both publicly available dataset ( FaceScape) and the newly constructed 3D face dataset.

\end{itemize}

In the rest of the paper, we introduce the proposed method in Section~\ref{proposedmethod} and experiments \& results in Section~\ref{expresults}. This is followed by a discussion about the different aspects of the proposed method in Section~\ref{sec:discussion}, followed by limitations \& potential future-works in Section~\ref{sec:limitations-future-works} and finally conclusions in Section~\ref{conclusion}.

\begin{figure*}[htp]
\begin{center}
\includegraphics[width=1.0\linewidth]{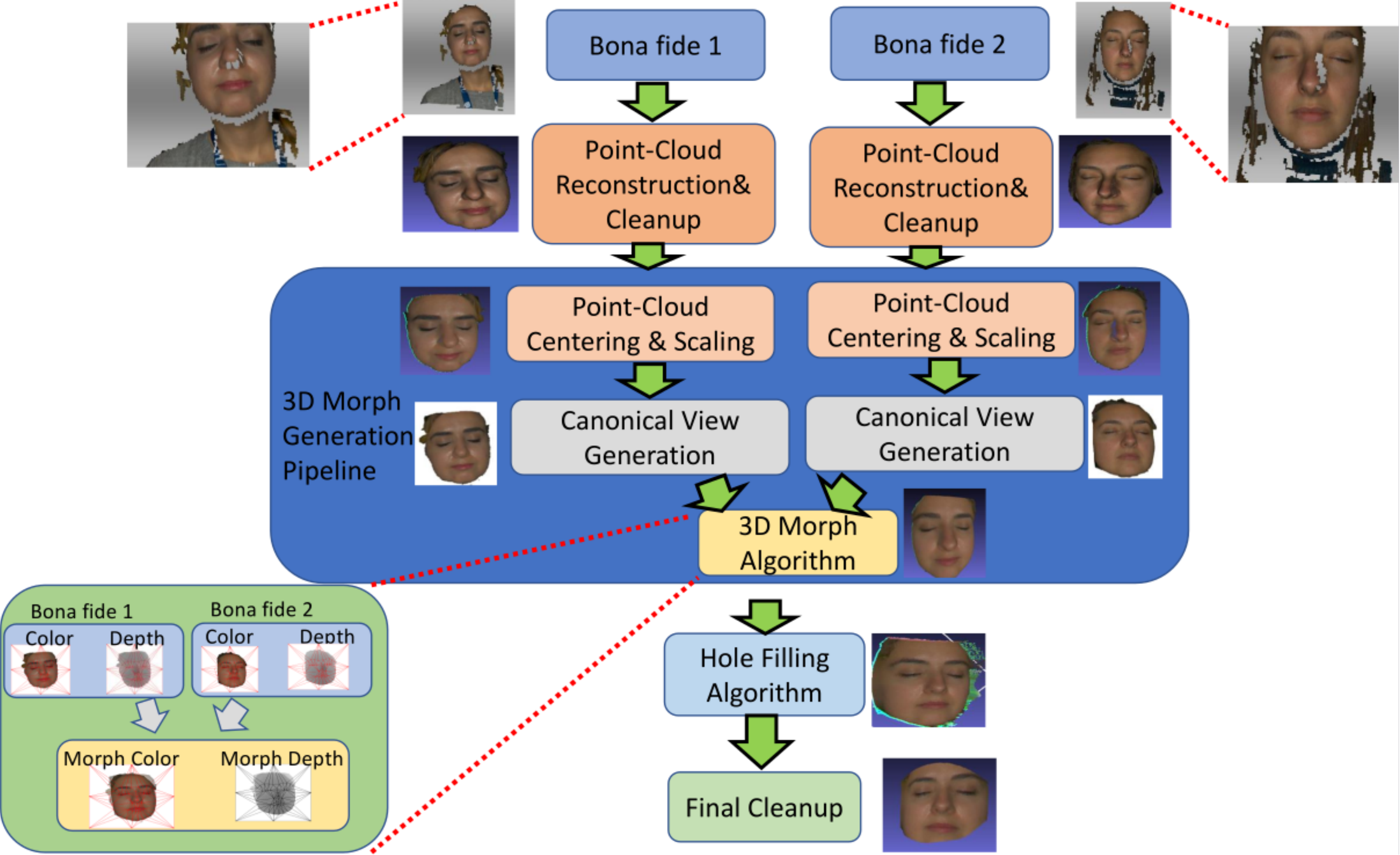}
\end{center}
   \caption{Block diagram of the proposed 3D face morphing generation technique}
\label{fig:blockDiagram}
\end{figure*}

\section{Proposed Method}\label{proposedmethod}

Figure \ref{fig:blockDiagram} shows a block diagram of the proposed 3D face-morphing generation framework based on the 3D point clouds. We are motivated to employ 3D point clouds over traditional 3D triangular meshes for two main reasons. The first is that connectivity information in a 3D triangular mesh leads to overhead storage, processing, management, and manipulation of the triangular meshes. Thus, 3D triangular meshes will significantly increase computing and memory, making them less suitable for low-computing devices. The second reason is that commodity scanning devices (for example, the Artec Sensor) can reproduce detailed colored point clouds that capture appearance and geometry. This allows us to generate high-quality 3D face morphing attacks.

However, 3D face morphing generation using point clouds introduces numerous challenges: (a)  Establishing a dense 3D correspondence between two different bona fide 3D point clouds to be morphed. Because 3D face point clouds from two different subjects are affected by various factors, such as differences in input point density, reliable detection of 3D facial key points, and estimation of affine/perspective warping,  (b)  locally affine deformation between two different 3D point clouds to be morphed is difficult to estimate~\cite{Yao_2020_CVPR,li08global,gmgp-global_reg-05}.  
(c) The misalignment of dense 3D correspondence between the two different 3D point clouds to be morphed increases with nonrigid deformation~\cite{besl1992method}.

A crucial part of 3D morphing using point clouds is reliable alignment before performing the morphing operation. Given the 3D face point clouds on the source and target faces, the point cloud registration can be defined as aligning a source point cloud to a target point cloud. The point cloud registration can be grouped into three broad categories \cite{Deng22_NonRigidSurvey_EGSTAR} namely 1) Deformation Field, 2) Extrinsic Methods and 3) Learning-based methods. Deformation-field-based techniques can be defined as the computation of deformation between two point clouds, which can be achieved either by assuming pointwise positions ~\cite{Liao09_DeformableSRGBD_ICCV} or by pointwise affine transformations~\cite{Yang19_GlobalRGBD_TIP}. Pointwise position variable methods are simplistic because they do not model deformations compared with pointwise affine transformations, which model local rotations. However, because local transformations must be stored and computed at a per-point level, this results in high computational and memory costs. This limitation was overcome by deformation-field-based methods using deformation graph embedding over the initial point set, which consisted of fewer nodes than the underlying point set~\cite{Li08_Global_CGF,Zampogiannis19_TopologyAware_TPAMI}. Extrinsic methods are based on optimizing an energy function to compute the point set correspondence, which usually includes an alignment term and regularization term~\cite{Li08_Global_CGF}. However, optimization-based methods compute deterministic modeling of the transformation. Probabilistic modeling of transformation was performed by Myronenko et al.~\cite{Myronenko10_CPD_TPAMI} in their algorithm, Coherent Point Drift (CPD), which assumes that the source points are centroids of an equally weighted Gaussian with an isotropic covariance matrix in the Gaussian Mixture Model (GMM). CPD consists of alignment and regularization terms for the transformation computation and performs non-rigid registration, but has memory and computation costs. However, the main limitation of optimization-based methods is that they produce good results when the input surfaces are close.
Furthermore, they require good initialization of the correspondences and the lack of these,  leading to convergence to local minima. This was overcome by learning-based data-driven methods of two types: (1) supervised methods and (2) unsupervised methods. Supervised methods require ground-truth data for training~\cite{Trappolini21_ShapeCorr_NeurIPS} but can work with varying point cloud density and underlying geometry. Unsupervised methods do not require ground-truth data and can be trained using a deformation module based on a CNN followed by an alignment module to compute the deformation~\cite{Zeng21_Corrnet3D_CVPR}.

However, the use of existing point cloud registration for this precise application of 3D face morphing point cloud generation poses challenges, such as {\bf{registration using the same individual}}, and point cloud registration has mainly focused on the non-rigid registration of two-point clouds from the same individual \cite{Deng22_NonRigidSurvey_EGSTAR}. This is primarily because high-quality registration is aimed at producing a globally consistent 3D mesh. Thus, the registration methods were not tested when two different point clouds were registered, as compared to those from the same individual. 
{\bf{Vertex accurate correspondence}}: 3D Face Morphing requires perfect vertex correspondence between the source and target point clouds, which is challenging and has not been extensively evaluated. 
{\bf{Low vertex count point clouds}}: Point cloud registration, especially when using learning-based methods, has network architectures based on point clouds with a low number of vertices (~1024). Thus, registering point clouds with many vertices (~ 75 K) has not been extensively evaluated and is therefore suitable for low-resolution face images. 
To address these challenges effectively, the proposed method consists of four stages: (1) point cloud reconstruction and cleanup, (2) 3D morph generation, (3) hole-filling algorithm, and (4) final cleanup. These steps are discussed in detail in the following subsections. 

\subsection{Point Cloud Reconstruction \& Cleanup}\label{stage1}
We captured a sequence of raw 3D scans by using the Artec Eva sensor~\cite{ArtecEvaSensor} from two data subjects to be morphed ($S_1$ and $S_2$). In this work, we consider the case of morphing two data subjects at a time because of its real-life applications, as demonstrated in several 2D face morphing studies  \cite{ferrara2014magic,venkatesh2021facesurvey}. We processed both $S_1$ and $S_2$ by performing a series of preprocessing operations such as noise filtering, texturing, and fusion of input depth maps to generate the corresponding point clouds $P_1$ and $P_2$. These operations were performed using Artec Eva Studio SDK filters together with the Meshlab filter \cite{meshlab}. The cleaned and processed point clouds are shown qualitatively in Figure~\ref{fig:blockDiagram}.

\subsection{3D Morph Generation Pipeline}\label{stage2}
In the next step, we process the point clouds $P_1$ and $P_2$ to generate a 3D face morphing point cloud by the following series of operations which are  discussed below: 
\subsubsection{\bf{Point-Cloud Centering \& Scaling}}\label{stage21}
First, we compute the minimum enclosing spheres using the algorithm from G{ \ "a}rtner et al.~\cite{Miniball} to obtain two bounding spheres with centers and radii ($C_1$,$r_1$), \& ($C_2$,$r_2$) corresponding to the point clouds $P_1$ and $P_2$ respectively. Note $P_1 = (v_1^{1},\ldots,v_1^{\textrm{n1}})$ where $v_1^{i}$ is the $i^{\textrm{th}}$ 3D vertex, $n1$ is the number of points in the point cloud $P_1$, and $P_2 = (v_2^{1},\ldots,v_2^{\textrm{n2}})$ where $v_2^i$ is the $i^{\textrm{th}}$ 3D vertex and $n2$ is the number of points in the point cloud $P_2$. We then subtract the sphere center $C_1$ from each 3D vertex of $P_1$ and repeat the same operation on  $P_2$ with $C_2$. Finally, the centered point clouds were scaled to the common radius, normalizing the 3D point clouds to a common scale. The resulting centered and scaled point clouds corresponding to $P_1$ and  $P_2$ are denoted as $PC_1$ and $PC_2$,  respectively. Figure~\ref{fig:blockDiagram} shows the qualitative results of this operation, which show the centered and scaled 3D point clouds.

\subsubsection{\bf{Canonical View Generation}}\label{stage22}
This step performs fine alignment by projecting the 3D face point clouds $PC_1$ and $PC_2$ onto the canonical (fixed) view. This step aims to keep the view and projection matrix identical to those of the 3D face point clouds  $PC_1$ and $PC_2$. We then projected $PC_1$ and $PC_2$ to generate 2D color images and depth maps using canonical view parameters. The generated 2D color images and depth maps are denoted  by ($I_1$,$D_1$) and ($I_2$,$D_2$), which correspond to the point clouds $PC_1$ and $PC_2$ respectively. In particular, we choose the canonical view for fine alignment because the traditional scheme of alignment, such as Iterative Closest Point (ICP)~\cite{besl1992method} does not provide a good alignment result when used on point clouds\cite{li08global}. This can be attributed to the limitations of the ICP to function when a locally affine/non-rigid deformation exists between the point clouds\cite{haehnel2003extension} 
The qualitative results of the canonical view transformation are shown in Figure~\ref{fig:blockDiagram}, which shows the aligned 2D color images and depth maps magnified in the inset image.

\begin{figure*}[h!]
     \centering
      \begin{tabular}[b]{c}
       \includegraphics[width=0.12\linewidth]{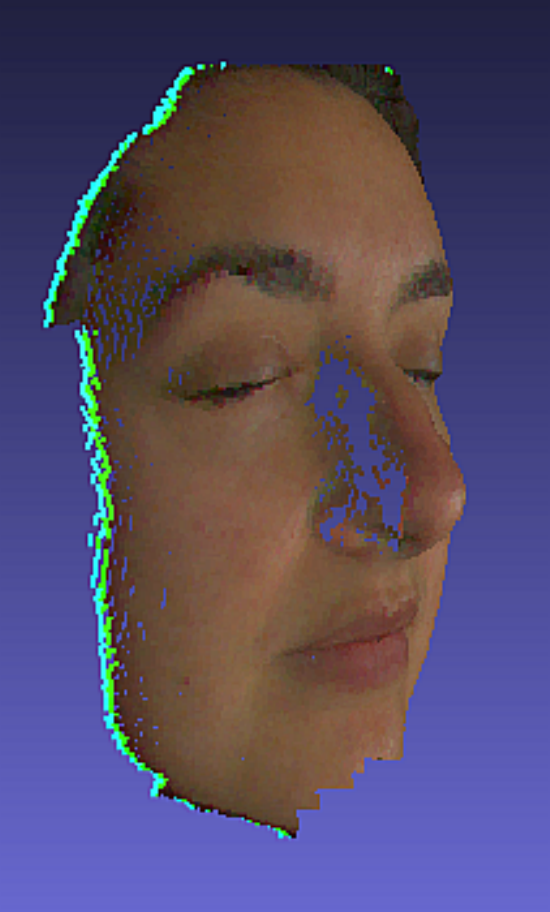}\\ 
    \small (a)
  \end{tabular} 
  \begin{tabular}[b]{c}
  \includegraphics[width=0.12\linewidth]{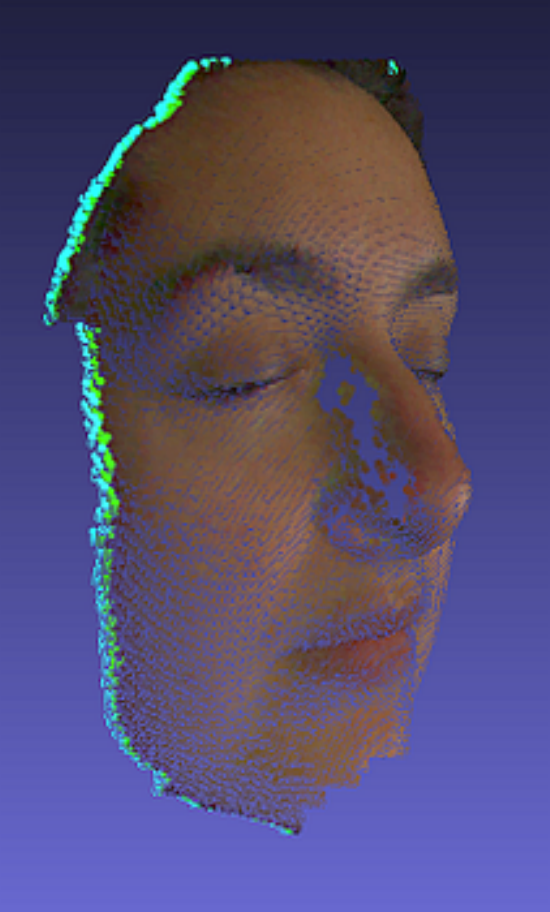}\\
    \small (b)
  \end{tabular} 
  \begin{tabular}[b]{c}
  \includegraphics[width=0.12\linewidth]{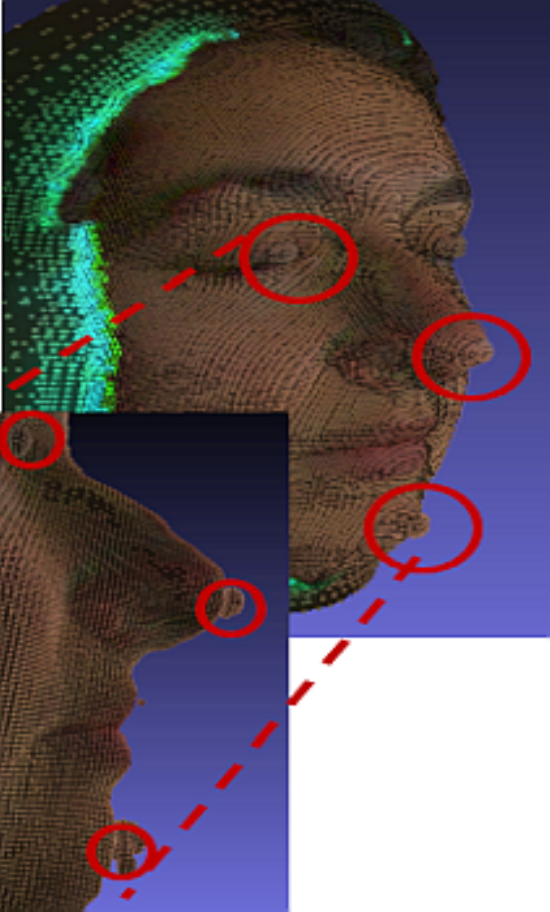}\\
    \small (c)
  \end{tabular}
  \begin{tabular}[b]{c}
  \includegraphics[width=0.12\linewidth]{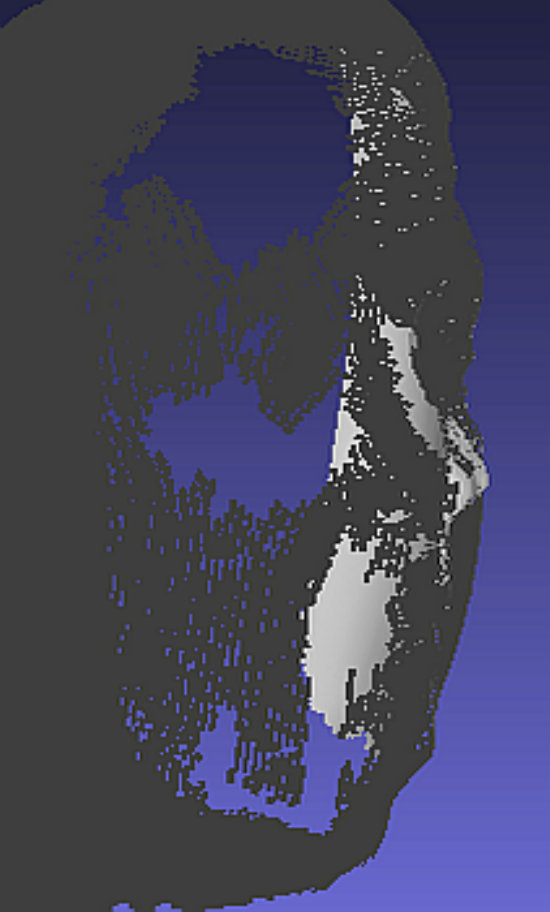}\\
    \small (d)
  \end{tabular}
  \begin{tabular}[b]{c}
  \includegraphics[width=0.12\linewidth]{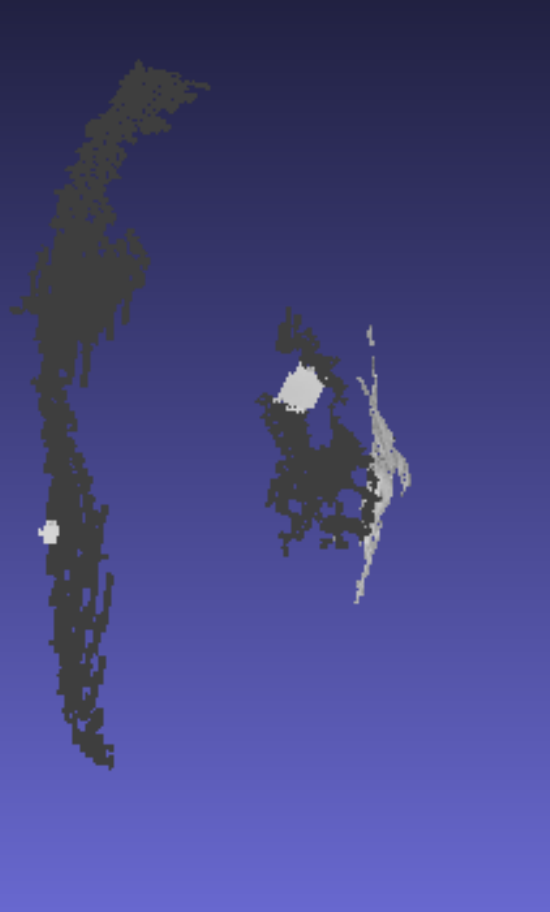}\\
    \small (e)
  \end{tabular}
  \begin{tabular}[b]{c}
  \includegraphics[width=0.12\linewidth]{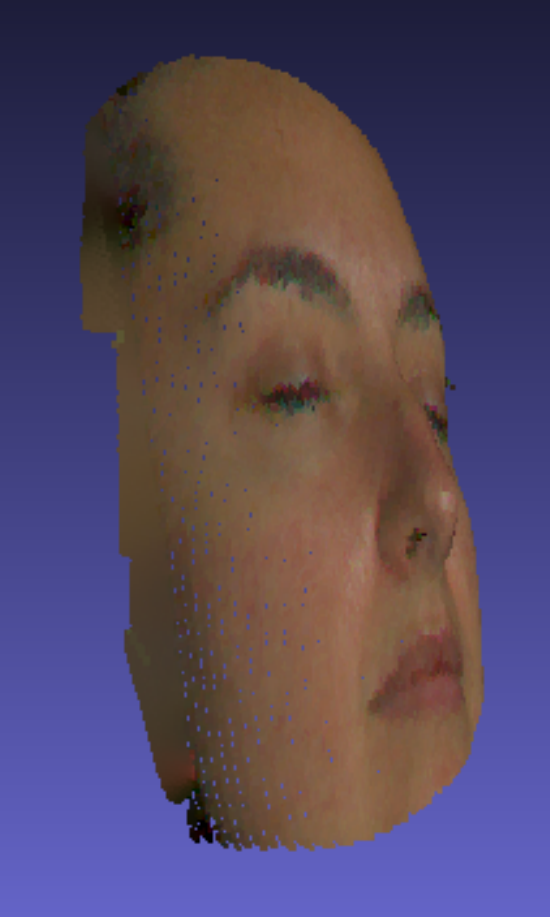}\\
   \small (f)
  \end{tabular}
     \caption{Qualitative results of the hole filling algorithms (a) Input Point Cloud with holes  (b) Point Cloud with Normals which has noise (c) Point Cloud with Screened Poisson Reconstruction~\cite{kazhdan2013screened} where artifacts are shown in the inset  (d) Point Cloud Reconstructed with APSS~\cite{apss}  (e) Point Cloud Reconstructed with RIMLS~\cite{rimls}  (f) Point Cloud Hole Filled using Proposed Method.}
     
     \label{fig:HFFFigure}
\end{figure*}

\begin{algorithm}[h!]
\caption{3D Face Morphing Algorithm}\label{3Dmorphalg} 
\hspace*{\algorithmicindent} \textbf{Input  ($I_1$, $I_2$, $D_1$, $D_2$, $CV$)} \\
    \hspace*{\algorithmicindent} \textbf{Output ($P_M$)} 
    \begin{algorithmic}[1]
         \STATE Detect Facial Keypoints on $K_1$ on $I_1$, and $K_2$ on $I_2$ using Dlib~\cite{dlib09}, and generate key-points of the\\ morph using Equation~\ref{eq1}.
         \STATE Perform Delaunay Triangulation on $K_M$\\  which is obtained by blending $K_1$ \\and $K_2$ using  Equation~\ref{eq1}.
         \STATE Estimate Affine Warping between corresponding triangles of $K_1$ \&  $K_M$ denoted as $w_1^M$, and for $K_2$ \&  $K_M$ denoted as $w_2^M$.
         \STATE Apply affine warping $w_1^M$ on $I_1$ to obtain $I_{1M}$, \\and on $D_1$ to obtain $D_{1M}$.
         \STATE Apply affine warping $w_2^M$ on $I_2$ to obtain $I_{2M}$, \\and on $D_2$ to obtain $D_{2M}$.
         \STATE Obtain morphed color image $I_M$ using the warped keypoints from the color images  $I_1$, and $I_2$ using Equation~\ref{eq1}, and morphed depth map $D_M$ using Equation~\ref{eq2}.
         \STATE Obtain the morphed point cloud by back-projecting\\  $I_M$, and $D_M$ to obtain the colored 3D point cloud $P_M$ \\with 3D coordinates ${\forall}i{\in}\{1,\cdots,n3\}(x_i,y_i,z_i)=(x_i,y_i,D_M(x_i,y_i))$ and color ${\forall}i{\in}\{1,\cdots,n3\}\textrm{Color}(x_i,y_i,z_i)=C_M(x_i,y_i))$ where $n3=\textrm{min}(n1,n2)$.  
         \end{algorithmic}
\end{algorithm}

\subsubsection{\bf{3D Morph Generation}}\label{stage23}
Given the 2D face color images ($I_1$,$I_2$) and  depth-maps ($D_1$,$D_2$) corresponding to $PC_1$, $PC_2$. We perform the morphing operation as explained in the Algorithm~\ref{3Dmorphalg}. The primary idea is to perform the morphing in 2D and back-project to 3D. The primary motivation for using a 2D morph generation method is to address the challenge of finding correspondence between  $PC_1$ and $PC_2$. The underlining idea is to perform the steps of morphing (facial landmark detection, Delaunay triangulation, \& warping) on  2D color images and re-use the same (facial landmark locations, triangulation, and warping) on the depth maps. In this work, we have used the blending (morphing) factor ($\alpha$) as $0.5$ as it is well demonstrated to be highly vulnerable in the earlier works on 2D face morphing~\cite{zhang2021mipgan}. The morphing is carried out as mentioned in the equation below:     
\begin{equation}\label{eq1}
\begin{split}
I_{M} = \alpha{\times}I_1(K_1^\prime)+(1-\alpha){\times}I_2(K_2^\prime)\\
K_1^\prime=w_1^M(K_1)\\
K_2^\prime=w_2^M(K_2)\\
K_M = \alpha{\times}K_{1}+(1-\alpha)*K_{2}
\end{split}
\end{equation}
where $\alpha$ is the blending factor, $K_1$ denotes 2D facial landmark locations corresponding to $I_1$, $K_2$ denotes 2D facial landmark locations corresponding to $I_2$, $K_M$ is generated by blending $K_1$, \& $K_2$, $w_1^M$ denotes the warping function from  $K_1$ to $K_M$,  $w_2^M$ denotes the warping function from  $K_2$ to $K_M$, and $I_M$ is the morphed 2D color image.
Similarly, the same operations are carried out on the depth maps as shown in the equation below:
\begin{equation}\label{eq2}
    D_{M} = \alpha{\times}D_1(K_1^\prime)+(1-\alpha){\times}D_2(K_2^\prime)\
\end{equation}
where $D_M$ is the morphed depth map.

In the next step, we back-project $I_M$, and $D_M$ to get the 3D face morphing point cloud $P_M=(v_M^{1},\ldots,v_M^{n3})$ where $n3=\textrm{min}(n1,n2)$ is the number of vertices. Note each 3D vertex is obtained using  ${i=1}^{n3}(x_i,y_i,z_i)=(x,y,D_M(x,y))$ and 
the qualitative results is shown in Figure~\ref{fig:blockDiagram}. However, generating the 3D face morphing will result in multiple holes due to a single canonical view. These holes are visible from other views. Therefore, we present a novel hole-filling algorithm to further improve the perceptual visual quality of the 3D face morphing.

\begin{figure*}
\centering
\begin{center}
\includegraphics[width=1.1\linewidth]{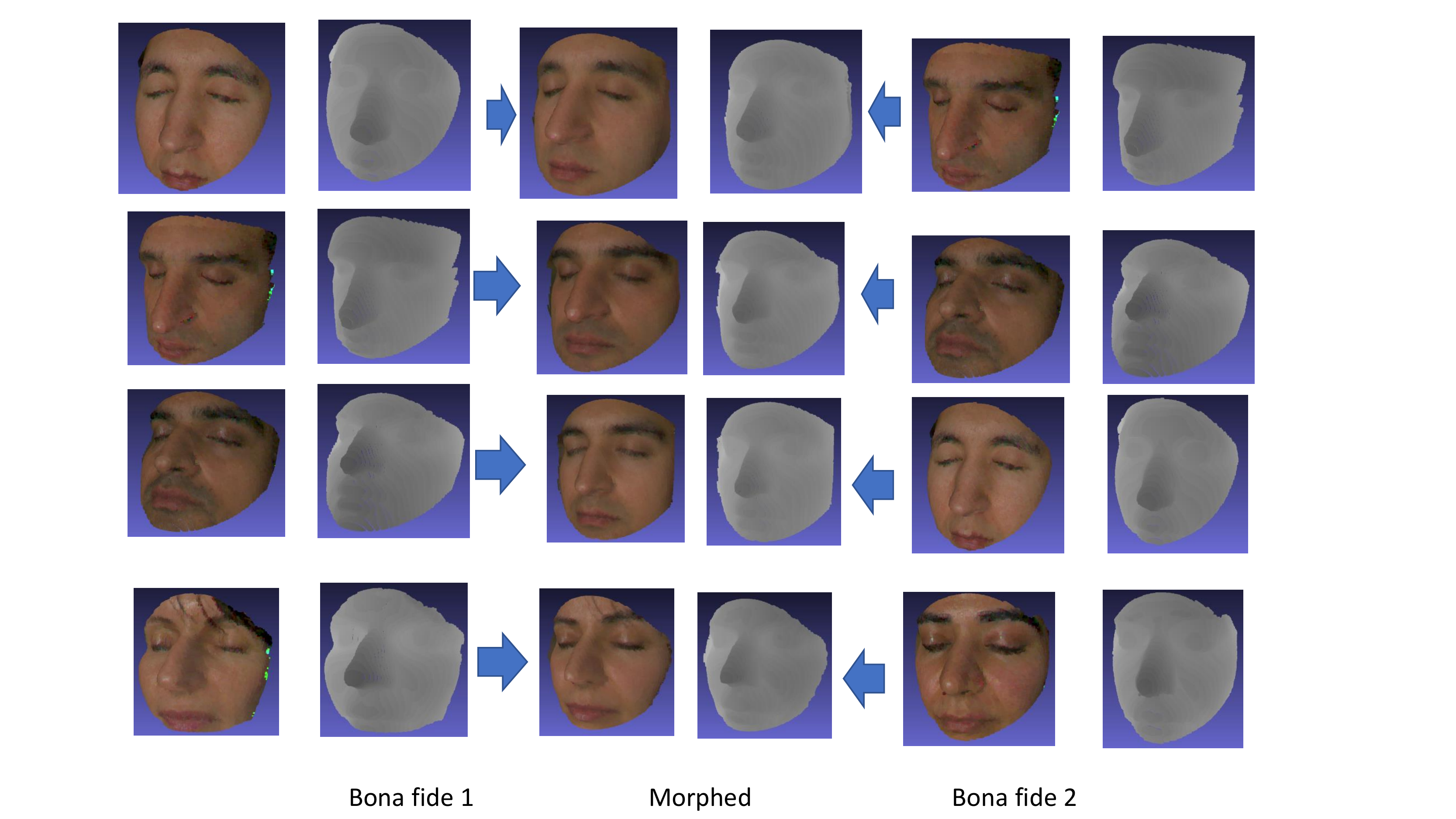}
\end{center}
   \caption{Illustration of 2D color image and depth maps for bona fide and morphs generated using the proposed method}
\label{fig:morphsamples}
\end{figure*}
\begin{algorithm}[h!]
\caption{Hole Filling Point Cloud}\label{holefillingalg} 
\hspace*{\algorithmicindent} \textbf{Input ($n4$-views)}\\
    \hspace*{\algorithmicindent} \textbf{Output ($C_{\textrm{hf}}$,$D_{\textrm{hf}}$,$P_{\textrm{hf}}$)} 
    \begin{algorithmic}[1] 
        \STATE Generate $n$ pairs of color-maps, and depth-maps  $\{(C_1,D_1),(C_2,D_2),\ldots,(C_j,D_j),\ldots,(C_{n4},D_{n4})\}$, translated from the canonical view.
        \FOR{$j \gets 1$ to $n4$}
         \STATE Perform Image In-painting~\cite{telea2004image} on $C_j$, and $D_j$.
         \STATE Perform Image Registration of  $C_j$ with the \\canonical view-point color-map $C_{\textrm{CV}}$ using\\ the following steps:
            \INDSTATE[2] Feature Computation using Oriented \\FAST and Rotated BRIEF (ORB) Descriptor~\cite{OrbFeature}.
            \INDSTATE[2] Brute-Force Matching of features using Hamming Distance.
            \INDSTATE[2] Homography computation using inlier\\ features.
            \INDSTATE[2] Perspectively warp the color and depth maps using computed homography.
        \ENDFOR{}
      \STATE Average all the registered color-maps ($C_{\textrm{hf}}$) and the depth-maps ($D_{\textrm{hf}}$).
      \STATE Back-Project the averaged color-map and\\ depth-map from 2D to 3D  to generate\\ hole-filled point cloud ($P_{\textrm{hf}}$) using the canonical view parameters.
         \end{algorithmic}
\end{algorithm}

\subsection{Hole Filling Algorithm}\label{stage5}
In this step, we propose a new hole-filling algorithm tailored to this specific 3D face morphing generation problem. Because the holes are visible from different views, filling the holes in these views is necessary to improve perceptual visual quality. {{Note that the holes are generated when the bona fide subject is looked at from a view different from the canonical camera, especially in high curvature regions such as the nose, as such areas are not completely visible from one canonical view.}} Therefore, we transform the 3D face-morphing point cloud $P_M$ multiple times independently to generate ${P_M^j}$ where $j=1{\ldots}n4$ and $n4$ is the number of transformations, and each transformation is a 3D translation \cite{foley1994introduction}. In this work, we empirically choose the number of 3D translations to $7$ to balance the computational cost and the visual quality achieved after hole filling. Using more 3D translations will significantly increase the computational cost and fail to improve visual quality. 
We tried the conventional approach of hole filling using 3D triangulation of 3D point clouds proposed in \cite{kazhdan2013screened},\cite{apss},\cite{rimls}. Figure~\ref{fig:HFFFigure} shows the qualitative results of the three different SOTA triangulation algorithms, which indicate unsatisfactory results. This is because the 3D orientation (3D normal) estimation indicates artifacts in the 3D triangulated mesh. Therefore, filling holes directly in the 3D point cloud is challenging because the underlying surface (manifold) is not known in advance. Errors in 3D orientation estimation make it difficult to employ conventional 3D hole-filling approaches.

This motivated us to devise a new approach for achieving effective hole filling. To this extent,  we project each point cloud $P_M^j$ onto the 2D face morphing color image ($C_j$) and its corresponding depth map ($D_j$). We fill the holes in $C_j$ \& $D_j$ using steps 2–9 described in Algorithm~\ref{holefillingalg}. Finally, we obtain the hole-filled 3D face morphing point cloud ($P_{\textrm{hf}}$), as indicated in Steps 10 and 11 in Algorithm~\ref{holefillingalg}. Figure \ref{fig:HFFFigure} (e) shows the qualitative results of the proposed hole-filling method, which indicates superior visual quality compared to the existing methods.

\subsection{Final Cleanup Algorithm}\label{stage6}
The final cleanup uses a clipping region outside a portion of the bounding sphere. The final result corresponding to the proposed 3D face morphing, a point cloud, is shown in Figure~\ref{fig:morphsamples} for an example data subjects \footnote{Supporting Video is available at \url{https://folk.ntnu.no/jagms/SupportingVideo.mp4}}.
The main advantages of the proposed method are as follows.
\begin{itemize}
\item The proposed method performs the alignment based on 2D facial key points, which preserves the identity in the generated 3D face morphing attack sample. 
\item The proposed method results in low computation and memory compared with existing 3D-3D techniques by overcoming the 3D registration.  
\item The proposed method results in a high vulnerability of FRS as the identity features are preserved for contributed data subjects used to generate the morphing attack. Therefore, the proposed method can cause high-quality 3D face-morphing attacks, resulting in vulnerability of both 2D and 3D face recognition systems. 
\item The proposed method can handle wide variation in the 3D pose.
\end{itemize}

\subsection{\bf{Qualitative and Quantitative Comparison of Proposed Method with SOTA}}
To illustrate the effectiveness of the proposed method, we selected a few SOTA methods based on nonrigid point cloud registration and methods generating a 3D face model from a  2D face image. Our current evaluation of SOTA for nonrigid point cloud registration (NRPCR) methods includes CPD by Myronenko et al.~\cite{Myronenko10_CPD_TPAMI} and  Corrnet3D by Zeng et al.~\cite{Zeng_2021_CVPR}. CPD is based on optimization and was the SOTA method for NRPCR earlier, whereas Corrnet3D is a more recent unsupervised deep-learning-based method for NRPCR. Furthermore, to evaluate the methods for generating a 3D face model from a 2D face image, we selected 3DMM by Blanz et al.~\cite{Blanz99_3DMM_SIGGRAPH} and a more recent deep-learning-based method, FLAME, by Li et al.~\cite{Li17_FLAME_SigAsia}. 3DMM introduced the concept of a morphable model, where parameters such as shape and texture can be controlled during 3D face synthesis. Furthermore, the 3DMM provided earlier SOTA results for 3D face generation from a 2D face image. FLAME enhances the quality of the generated 3D face model from a 2D face image by using more controllable parameters such as pose, expression, shape, and texture during the 3D face synthesis process. 

\subsubsection{\bf{Qualitative Comparison and Analysis}}
\begin{figure*}[htp]
\begin{center}
\includegraphics[width=0.9\linewidth]{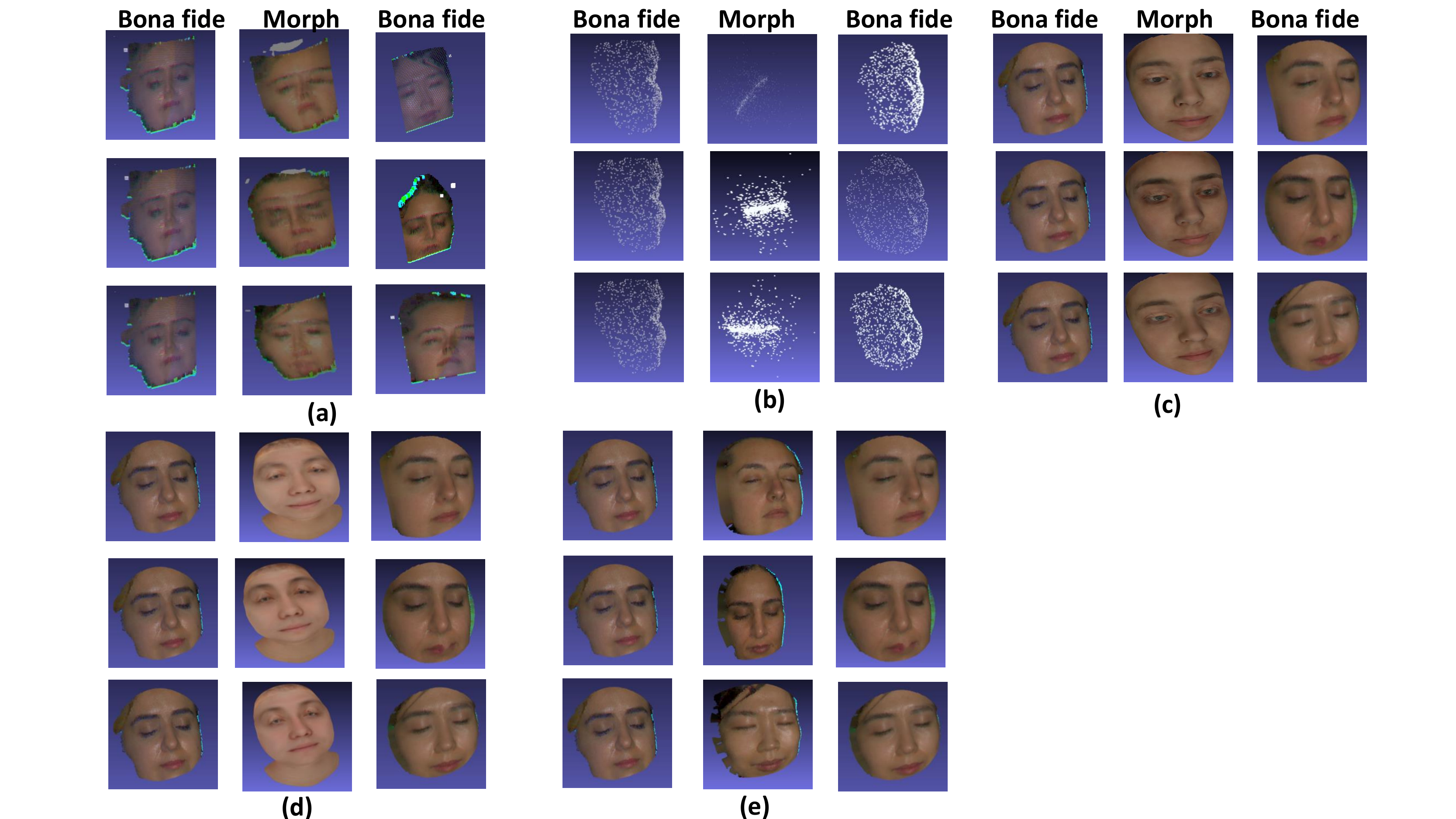}
\end{center}
   \caption{{Illustration of the SOTA Comparison showing Bona fide and Morphs generated using (a) CPD~\cite{Myronenko10_CPD_TPAMI}, (b) Corrnet3D~\cite{Zeng_2021_CVPR}, (c) 3DMM~\cite{Blanz99_3DMM_SIGGRAPH} (d) FLAME~\cite{Li17_FLAME_SigAsia}, (e) Proposed Method. Note that both 3DMM and FLAME need a single image as input, and in the current evaluation, we pass a 2D rendering generated using the proposed method. Note that the proposed method shows high-quality rendering and identity features of the 2D face morphing image.}}
\label{fig:sotaComparisonQualitative}
\end{figure*}

The results of qualitative comparison with SOTA are shown in Figure~\ref{fig:sotaComparisonQualitative} and the quantitative vulnerability computed using MMPMR \cite{MMPMRUlrich} and FMMPMR \cite{FMMPMRSushma} (refer  Section \ref{sec:vul} for the definition of these metrics) is indicated in the Table \ref{table:smalldatasetfmmpmr}.  
It can be noticed from Figure~\ref{fig:sotaComparisonQualitative} that SOTA methods don't contain identity features of the 3D face morphing model to a large extent. However, CPD does contain the identity features of the 3D face morphing model but fails on the alignment of the two input point clouds, which results in double features such as eyebrows. Corrnet3D produces lower-quality results, which can be attributed to the fact that the authors have not yet focused exclusively on face registration.

Further, 3DMM and FLAME generate a 3D face model from a 2D face image. Thus, we passed the rendering (2D face image) of the 3D face morphing model as an input. However, these methods fail to preserve the identity features during the 3D face model generation, as seen from Figure~\ref{fig:sotaComparisonQualitative}. The generated 3D model has a low resemblance to the identity features of the face morphing image.

\begin{table}[!ht]
    \centering
   \caption{{{Vulnerability of SOTA on Comparison Dataset}}}
    \resizebox{\linewidth}{!}{
    {{
   \begin{tabular}{|c|c|c|c|c|}
    \hline
    
        \textbf{Feature} & \textbf{3DMM}~\cite{Blanz99_3DMM_SIGGRAPH} & \textbf{FLAME}~\cite{Li17_FLAME_SigAsia} & \textbf{CPD}~\cite{Myronenko10_CPD_TPAMI} & \textbf{Proposed} \\ \hline
        PointNet++~\cite{qi2017pointnet++} & 0 & 0 & 0 & 100\% \\ \hline
        LED3D~\cite{Led3D2019} & 66.67\% & 0 & 0 & 100\% \\ \hline
    \end{tabular}}}}
    \label{table:smalldatasetfmmpmr}
\end{table}


\begin{figure*}[htp]
\begin{center}
\includegraphics[width=0.45\linewidth]{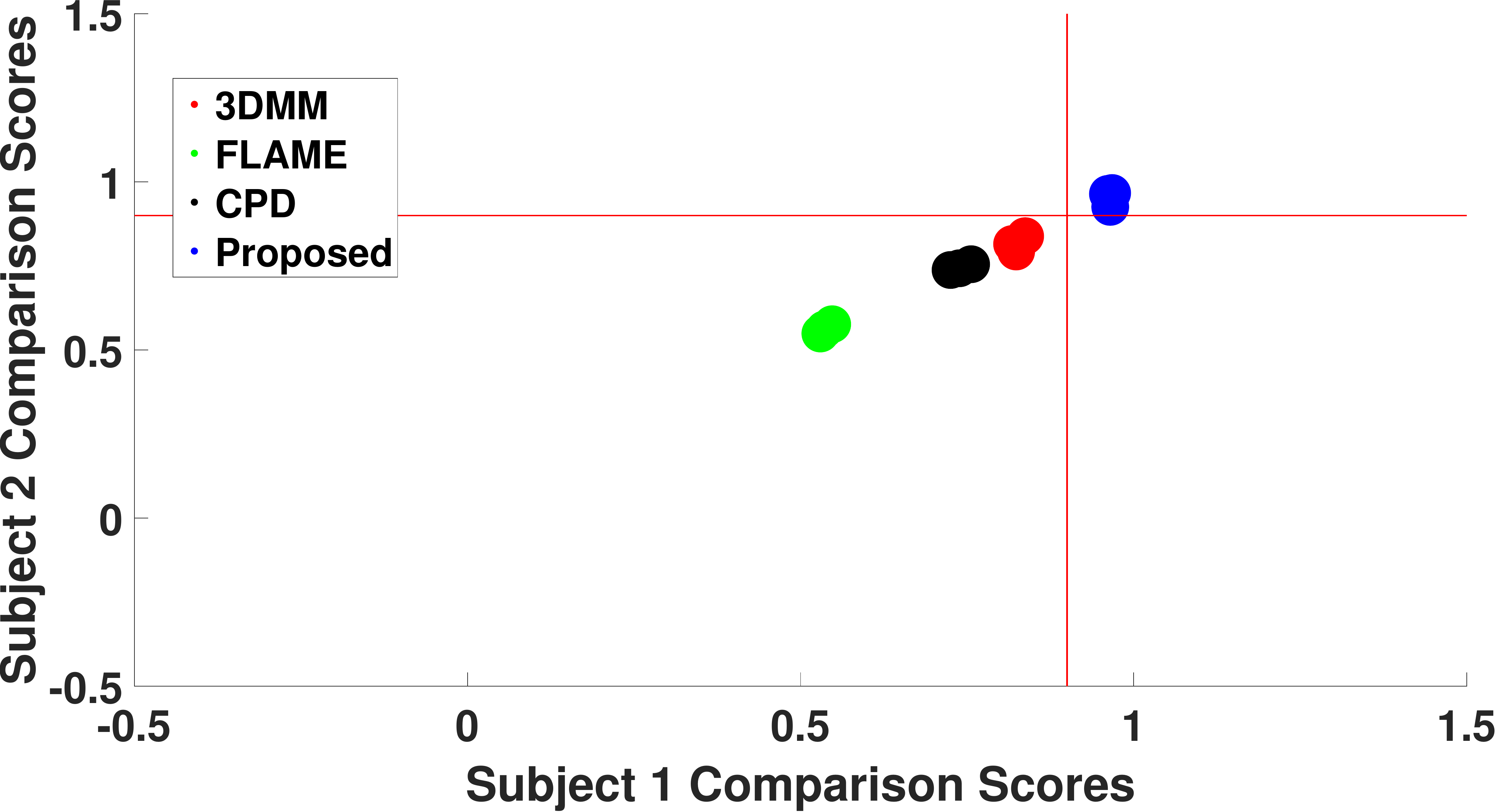}
\includegraphics[width=0.45\linewidth]{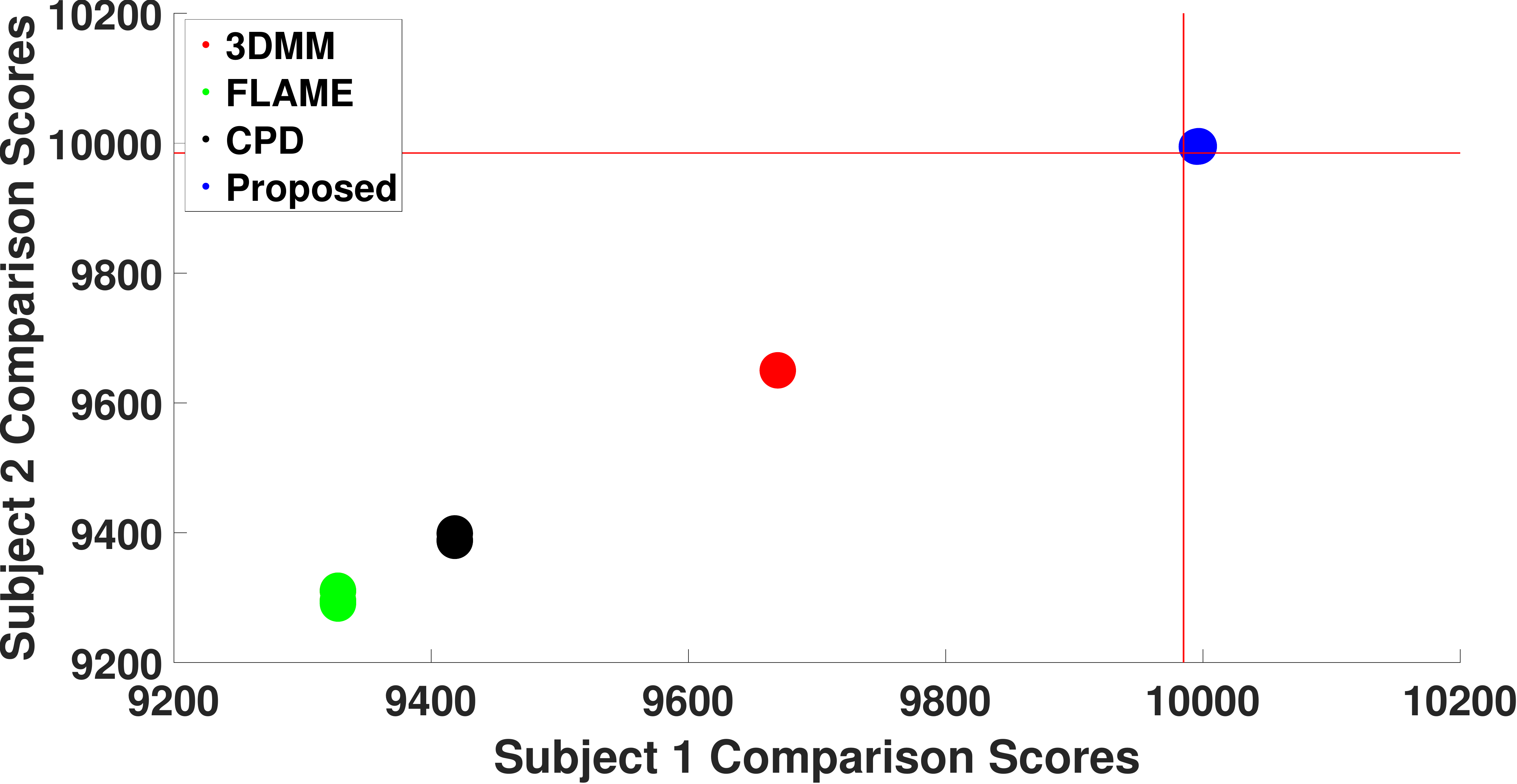}\\
(a) \hspace{80mm} (b)
\end{center}
   \caption{{Illustration showing scatter plot of Comparison scores using Bona fide and Morphs generated using Proposed Method
   (a) LED3D~\cite{Led3D2019} and (b) Pointnet++~\cite{qi2017pointnet++} based where SOTA algorithms are 3DMM~\cite{Blanz99_3DMM_SIGGRAPH}, FLAME~\cite{Li17_FLAME_SigAsia}, CPD~\cite{Myronenko10_CPD_TPAMI}}}
\label{fig:sotaComparisonQuantitative}
\end{figure*}
\begin{figure*}[htp]
\begin{center}
\includegraphics[width=0.9\linewidth]{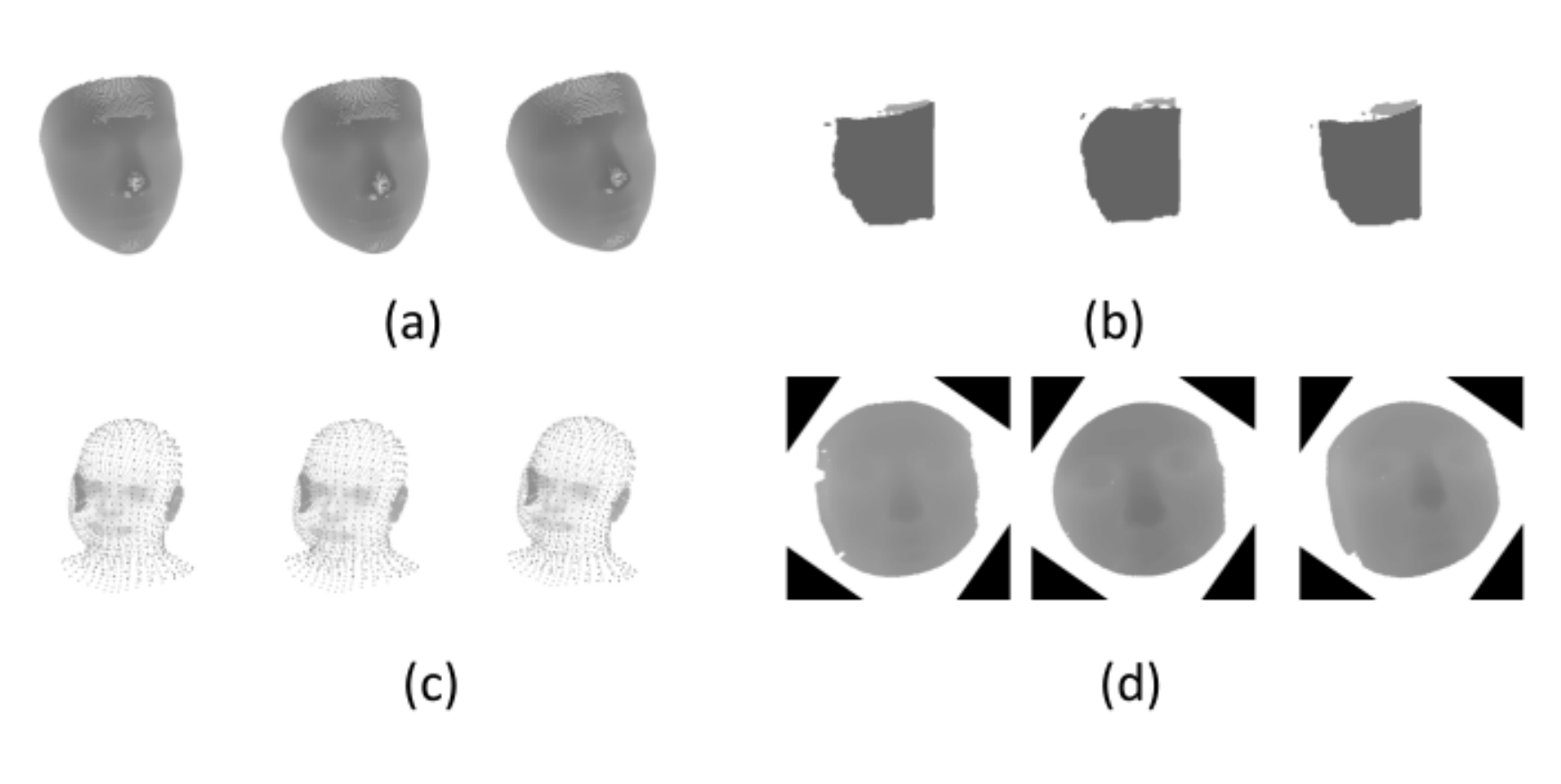}
\end{center}
   \caption{Illustration showing depth maps using SOTA and proposed method
   (a) 3DMM~\cite{Blanz99_3DMM_SIGGRAPH}, (b) CPD~\cite{Myronenko10_CPD_TPAMI}, (c) FLAME~\cite{Li17_FLAME_SigAsia} and (d) Proposed Method.}
\label{fig:depthmapsComparison}
\end{figure*}

\begin{figure}
        \centering
    \includegraphics[width=\linewidth]{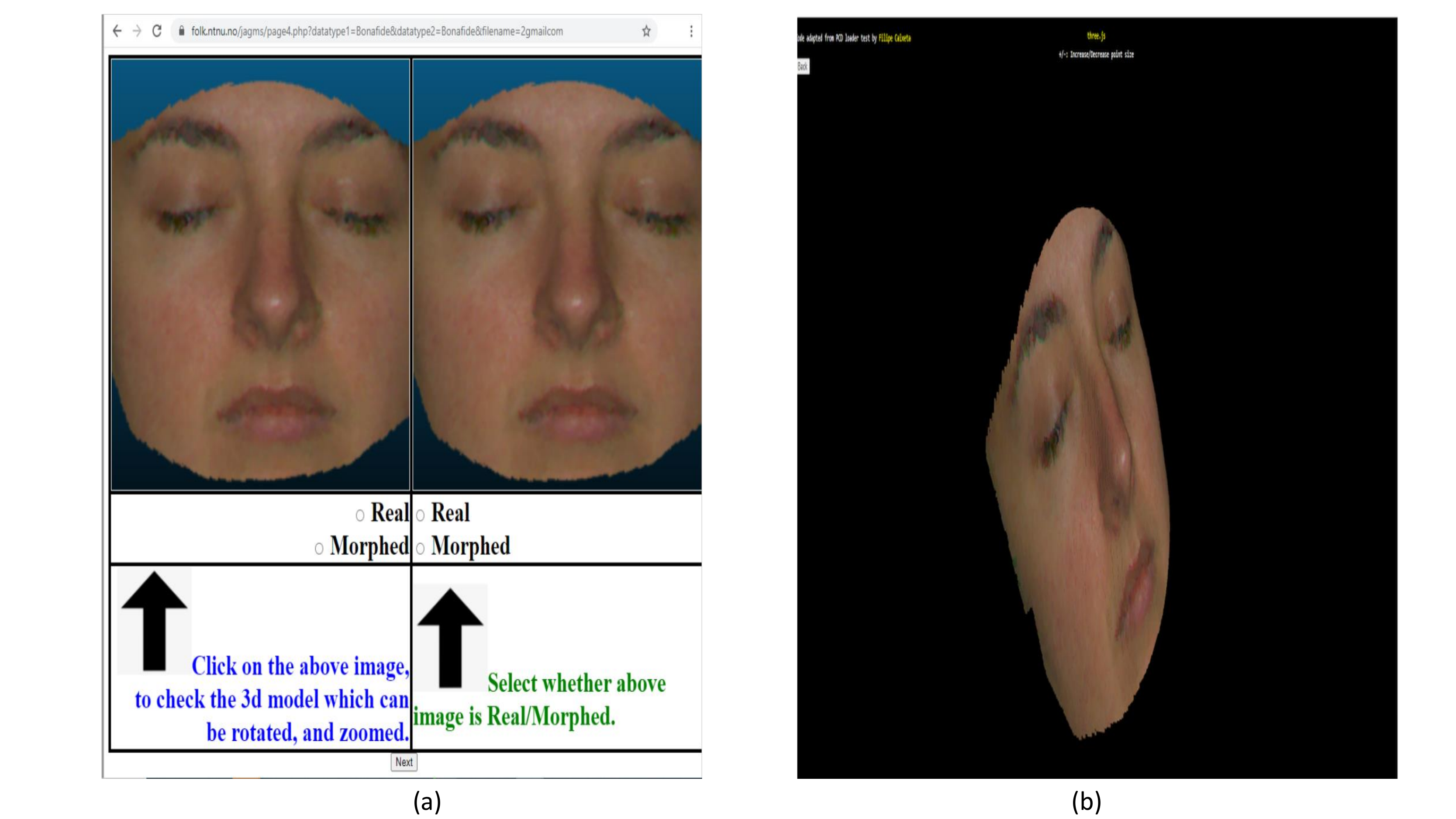}
\caption{Screenshots from the GUI of human observer web page  (a) Full Page Screenshot, and (b) Screenshot of 3D model page.}\label{fig:Screenshots}
\end{figure}

\begin{figure}
\begin{center}
\includegraphics[height=0.6\linewidth]{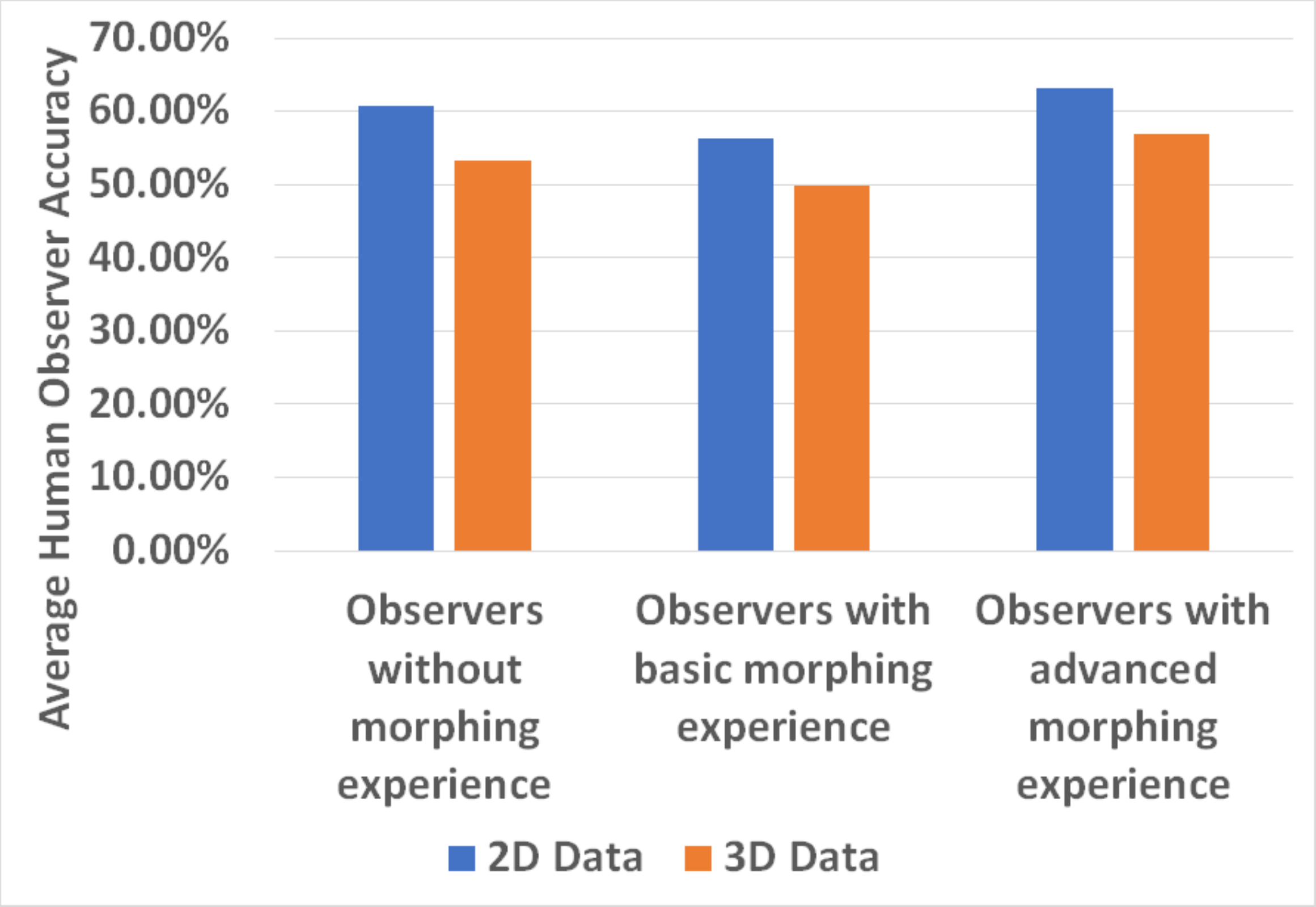}
\end{center}
   \caption{Illustration of average accuracy of human observer study, note that 2D accuracy is always higher than 3D.}
\label{fig:morphingsurvey}
\end{figure}

\subsubsection{\bf{Quantitative Comparison and Analysis}}
The results of the quantitative comparison are shown in Figure~\ref{fig:sotaComparisonQuantitative}, where we have evaluated two 3D point feature extraction methods, namely LED3D~\cite{Led3D2019} and Pointnet++~\cite{qi2017pointnet++}. However, it can be seen that 3D comparison results in low values for SOTA compared to the proposed method. This can be attributed to the low-resolution of the identity-specific depth generation by the SOTA, which is also shown in Figure~\ref{fig:depthmapsComparison}.

A sample implementation of the proposed method is available at \footnote{Proposed Method Implementation \url{https://github.com/jagmohaniiit/3DFaceMorph}}.

\section{Experiments and Results}
\label{expresults}
In this section, we present the discussion on extensive experiments carried out on the newly acquired 3D face dataset. We discuss the quantitative results of the various experiments, including vulnerability study on automatic FRS and human observer study, quantitative quality estimation based on color and geometry of the generated 3D face morphing models and automatic detection of 3D MAD attacks.  

\begin{table}[htp]
    \centering
     \caption{Statistics of newly collected 3D  Morphing Dataset (3DMD)}
    \resizebox{.8\linewidth}{!}{
    \begin{tabular}{|c|c|c|} 
    \hline
    \multicolumn{3}{|c|}{{\bf{3D face Bona fide}}} \\  \hline \hline
    \textbf{Total Data Subjects} & Males & Females\\ \hline
    41 & 28 & 13 \\\hline
    Total 3D samples & Males & Females \\ \hline \hline
    330    & 224 & 106 \\ \hline
    \multicolumn{3}{|c|}{{\bf{3D face Morphs}}} \\  \hline \hline
    \textbf{Total 3D Morphs} & Males & Females\\ \hline
    345 & 278 & 67 \\ \hline
           \end{tabular}}
            
    \label{table:datasetdetails}
  
\end{table}

\subsection{3D Face Data Collection}


In this study, we constructed a new 3D face dataset using the Artec Eva  3D scanner \cite{ArtecEvaSensor}. Data collection was conducted in an indoor lighting environment. The subjects were asked to sit on the chair by closing their eyes to avoid strong reflection of light from the 3D scanner. The 3D scanner was moved vertically to capture the 3D sequence. 

Artec Studio Professional 14 was used for 3D data collection and processing. We collected 3D facial data from 41 subjects, including 28 males and 13 females. We captured nine–ten samples for each data subject in three different sessions over three days. The statistics of the whole 3D face dataset are summarized in Table~\ref{table:datasetdetails}. We name our newly collected dataset the 3D Morphing Dataset (3DMD).

We may have used existing 3D face datasets such as FRGC~\cite{frgc} and BU-3DFE~\cite{bu3dfe}. However,  the FRGC dataset provides a single depth map and color image. Thus, a high-quality point cloud cannot be generated. Furthermore, the dataset has a few misaligned color images and depth maps \cite{frgcreason} which results in low-quality 3D morphing generation. The BU-3DFE~\cite{bu3dfe} dataset provides 3D models, but these are perfectly registered and the capture conditions are identical for all subjects. This does not model the real-world scenario of capturing 3D point clouds with changes in the capture conditions that could occur during data collection. The quality of our 3D face dataset has a much higher number of 3D vertices between 35950 \& 121088 for the inner face compared to previous methods~\cite{egger20203d}. These factors motivated us to generate a new 3D face dataset to enable high-quality 3D face morphing generation suitable for ID control scenarios.


\begin{figure*}
        \centering
        \begin{subfigure}[b]{0.25\textwidth}
                \centering
                \includegraphics[width=\textwidth]{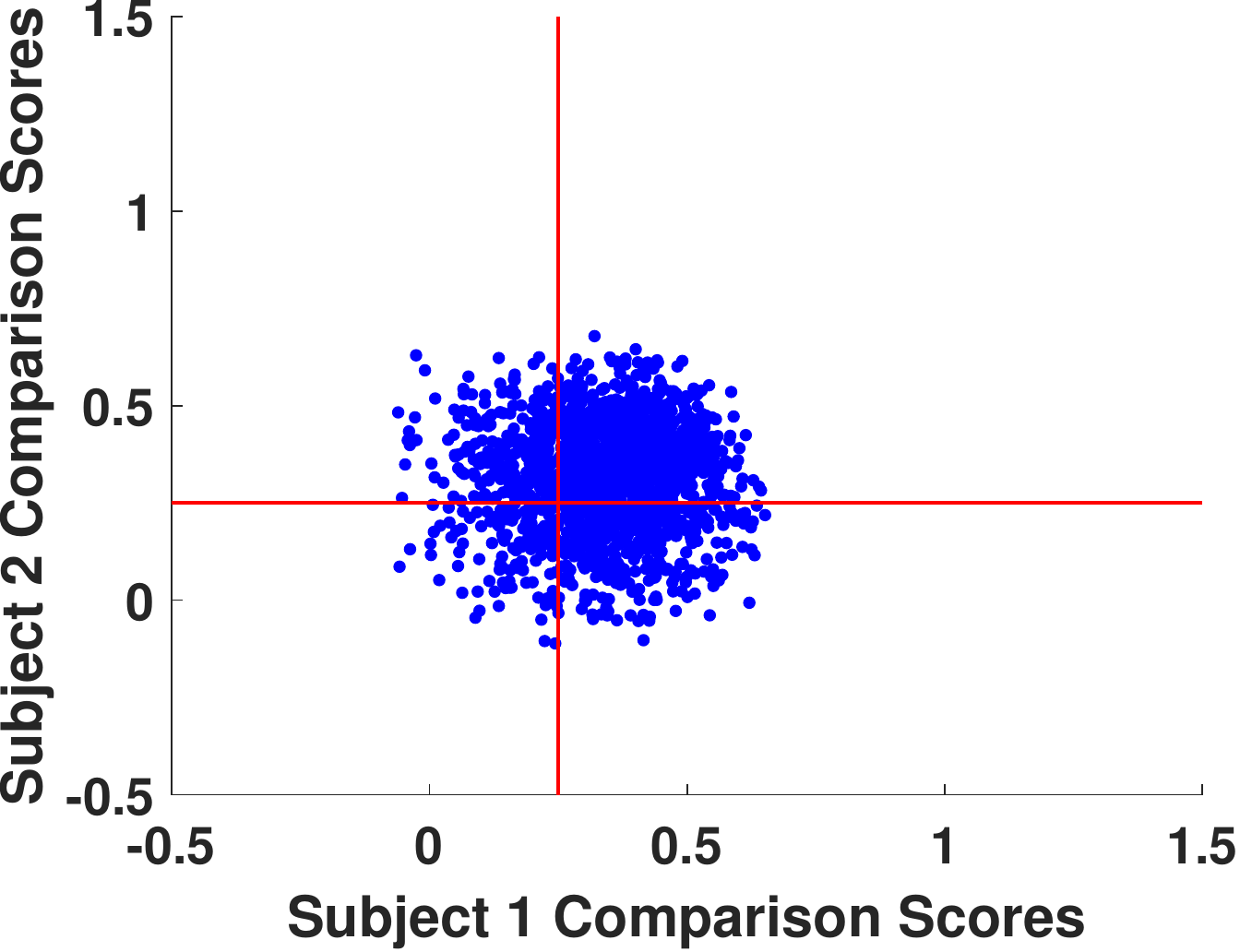}
                \caption{}
                \label{fig:gull1}
        \end{subfigure}%
        \begin{subfigure}[b]{0.25\textwidth}
                \centering
                \includegraphics[width=\textwidth]{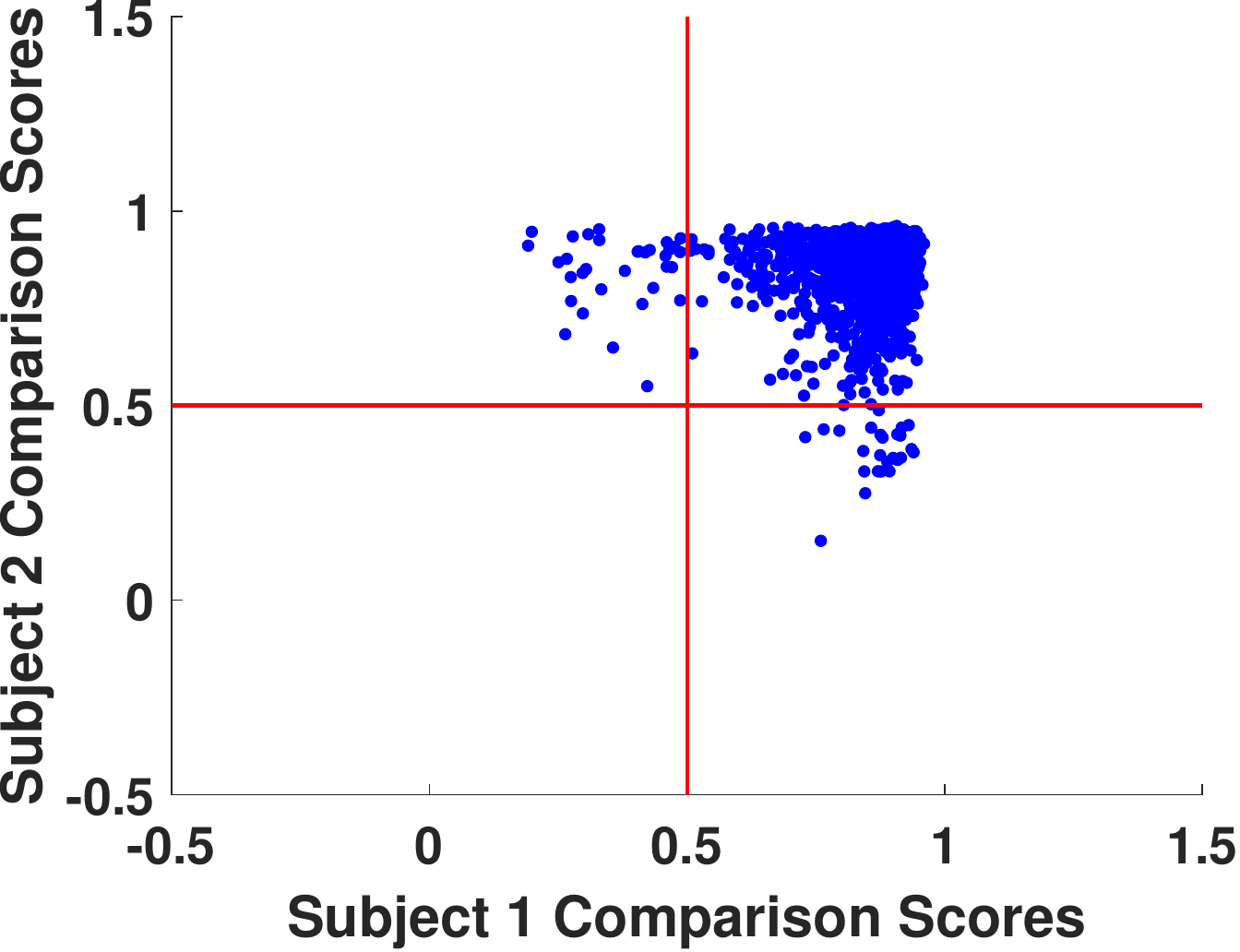}
                \caption{}
                \label{fig:gull2}
        \end{subfigure}%
         \begin{subfigure}[b]{0.25\textwidth}
                \centering
                \includegraphics[width=\textwidth]{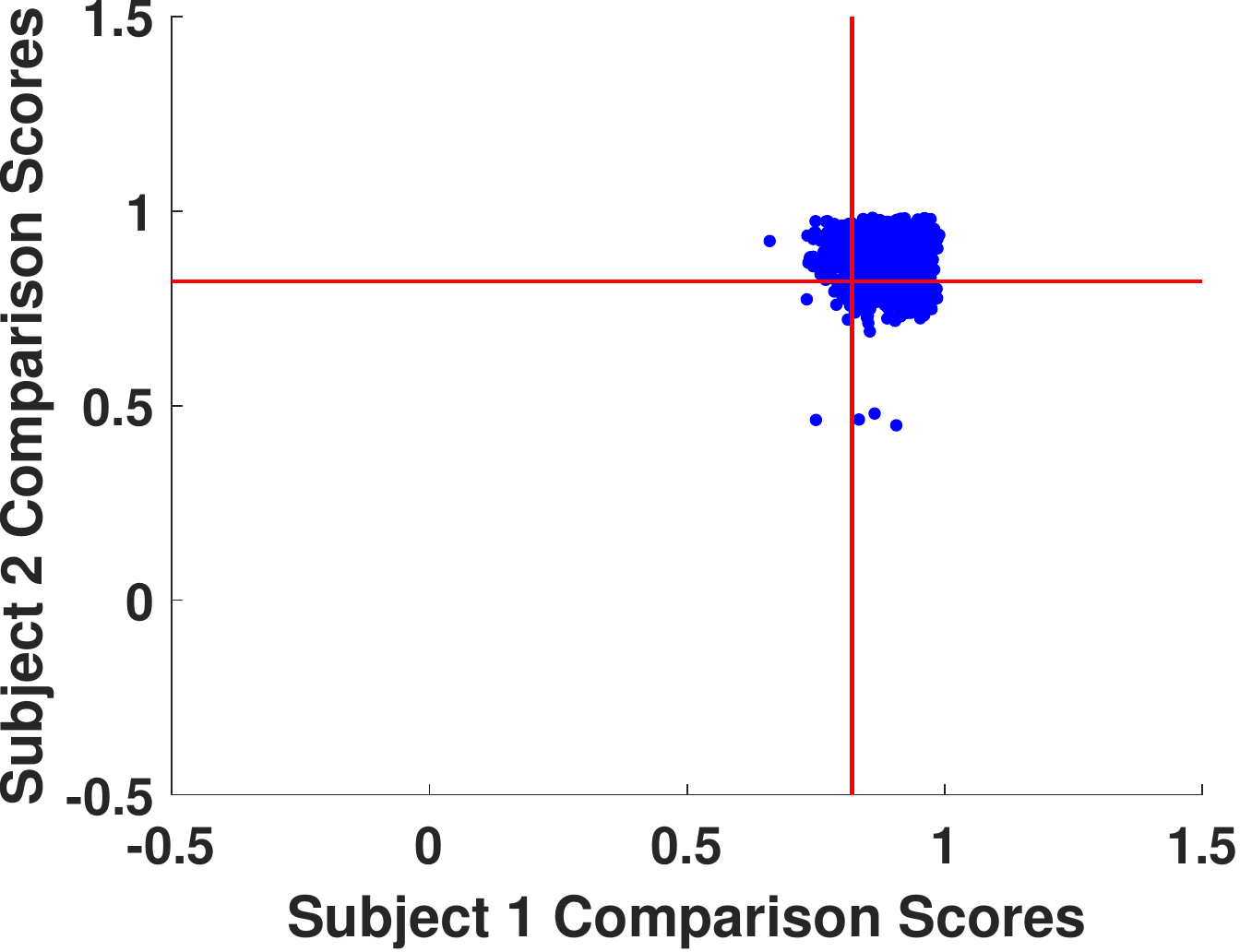}
                \caption{}
                \label{fig:gull3}
        \end{subfigure}%
             \begin{subfigure}[b]{0.25\textwidth}
                \centering
                \includegraphics[width=\textwidth]{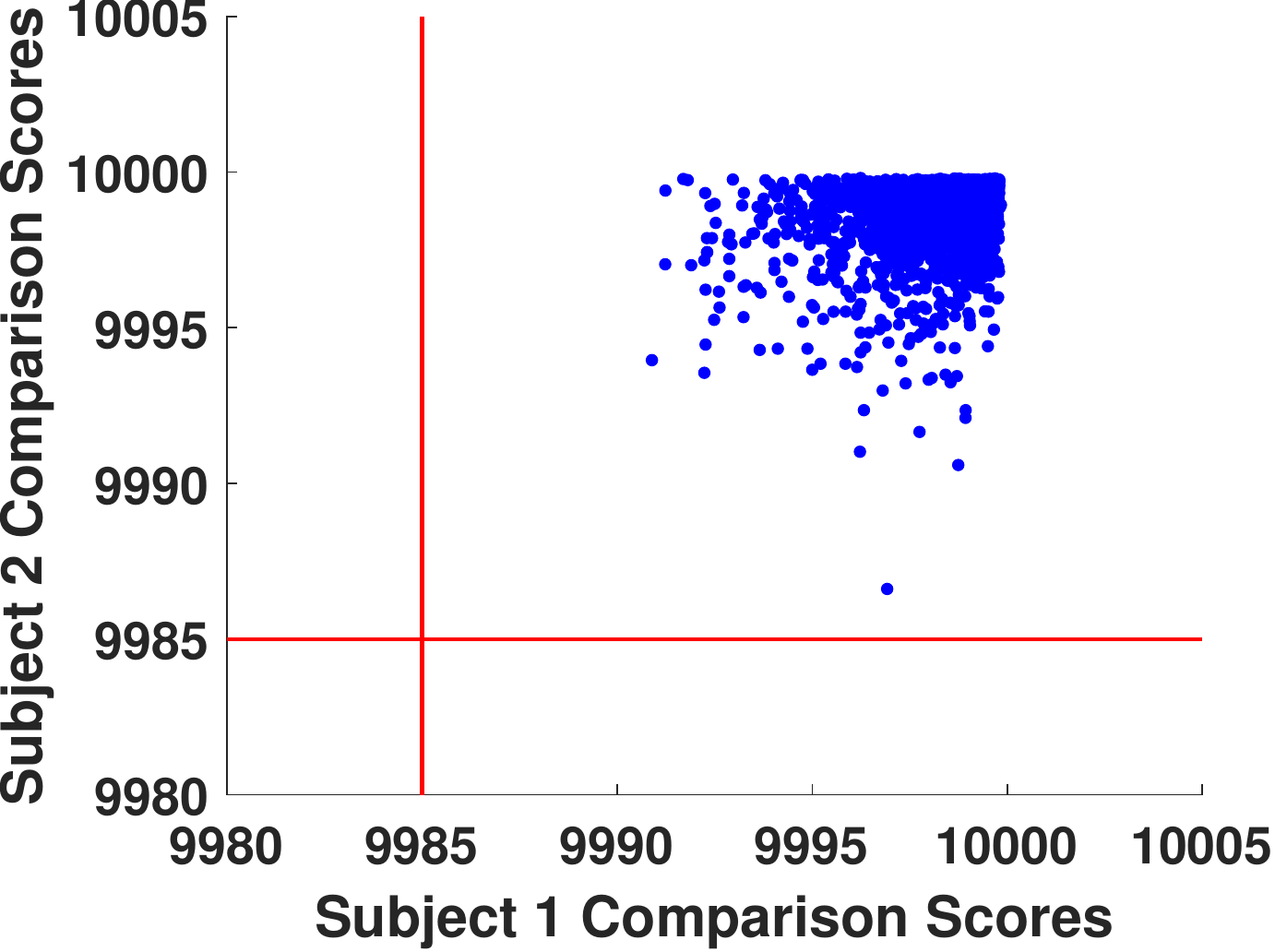}
                \caption{}
                \label{fig:gull4}
        \end{subfigure}%
               \caption{Vulnerability Plots using 2D \& 3D FRS on 3D Morphing dataset (3FMD) (a) 2D face FRS using Arcface~\cite{deng2019arcface}, (b) 2D face FRS using COTS, and (c) 3D face FRS  using Led3D~\cite{Led3D2019}, and (d) 3D face FRS  using  Pointnet++~\cite{qi2017pointnet++}}\label{figRFFiguresProposed}
\end{figure*}

\begin{figure*}
        \centering
        \begin{subfigure}[b]{0.25\textwidth}
                \centering
                \includegraphics[width=\textwidth]{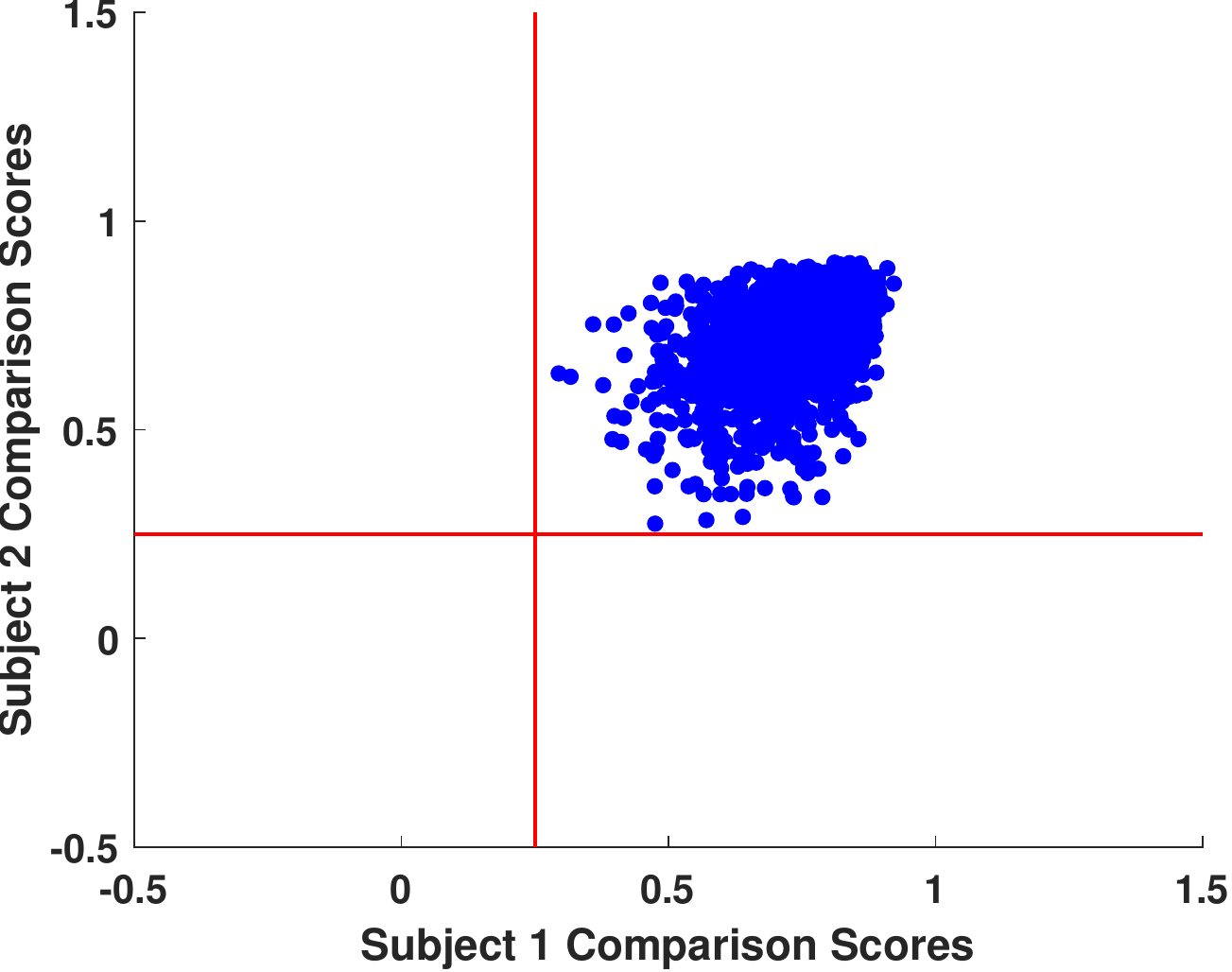}
                \caption{}
                \label{fig:gull5}
        \end{subfigure}%
        \begin{subfigure}[b]{0.25\textwidth}
                \centering
                \includegraphics[width=\textwidth]{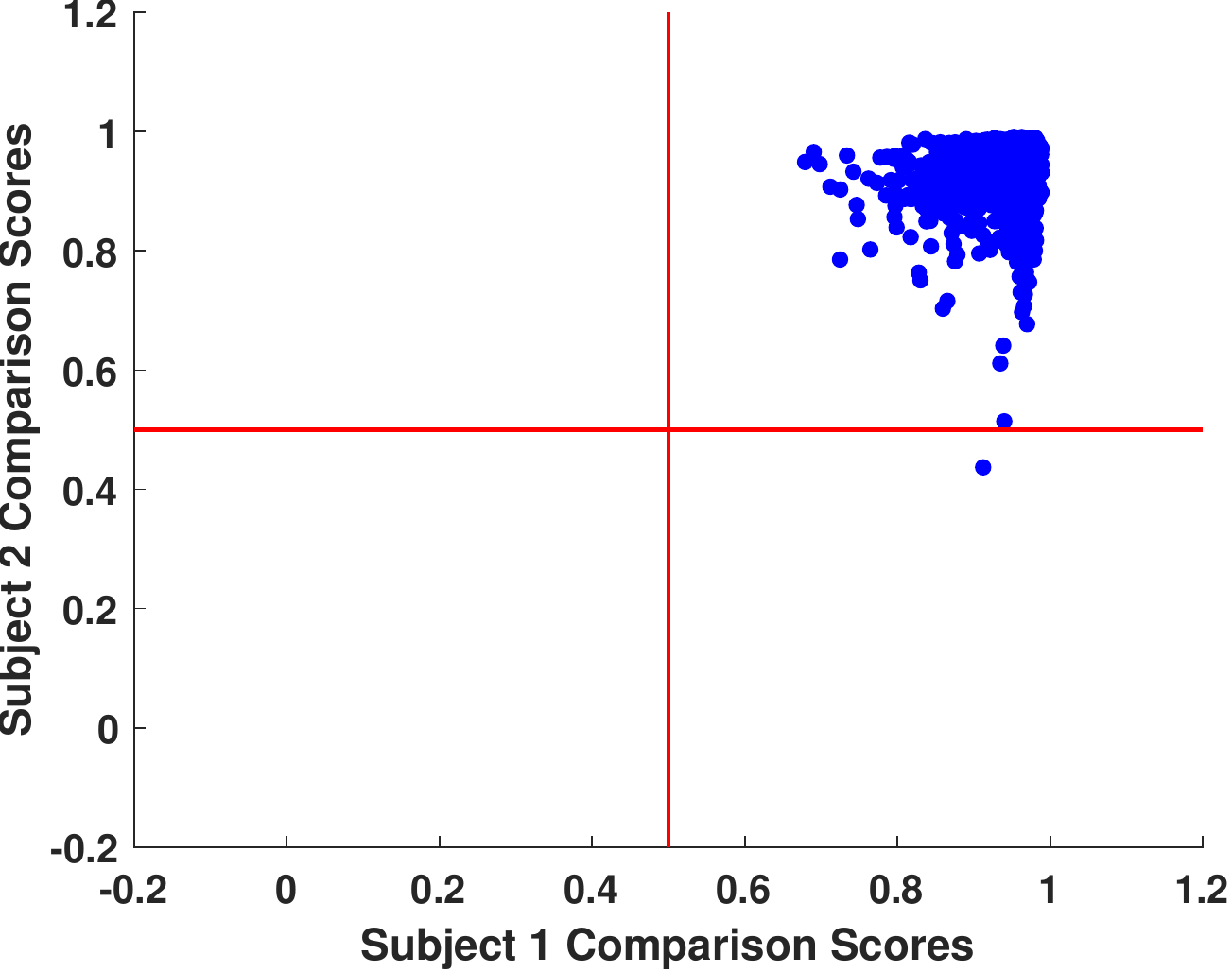}
                \caption{}
                \label{fig:gull6}
        \end{subfigure}%
         \begin{subfigure}[b]{0.25\textwidth}
                \centering
                \includegraphics[width=\textwidth]{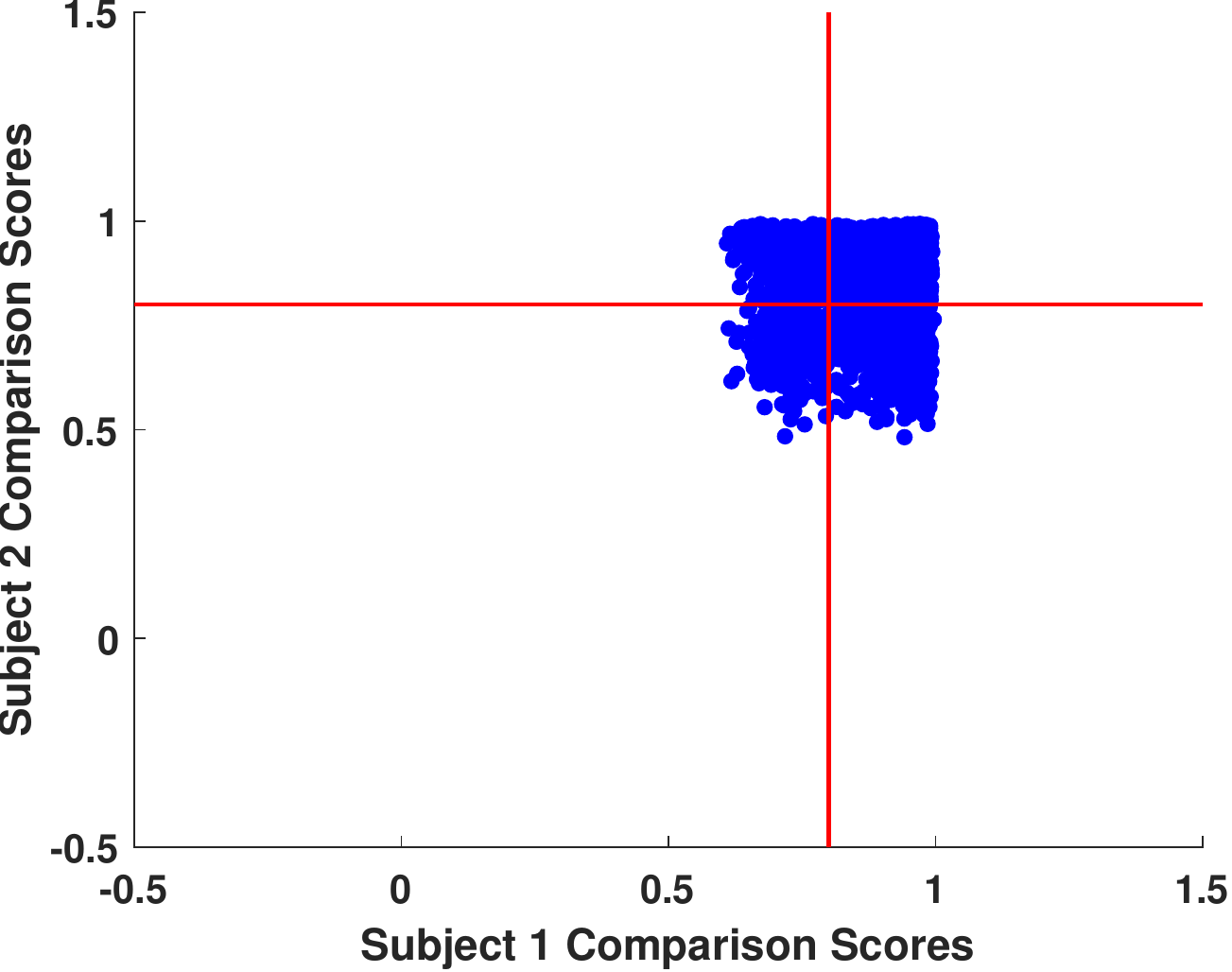}
                \caption{}
                \label{fig:gull7}
        \end{subfigure}%
             \begin{subfigure}[b]{0.25\textwidth}
                \centering
                \includegraphics[width=\textwidth]{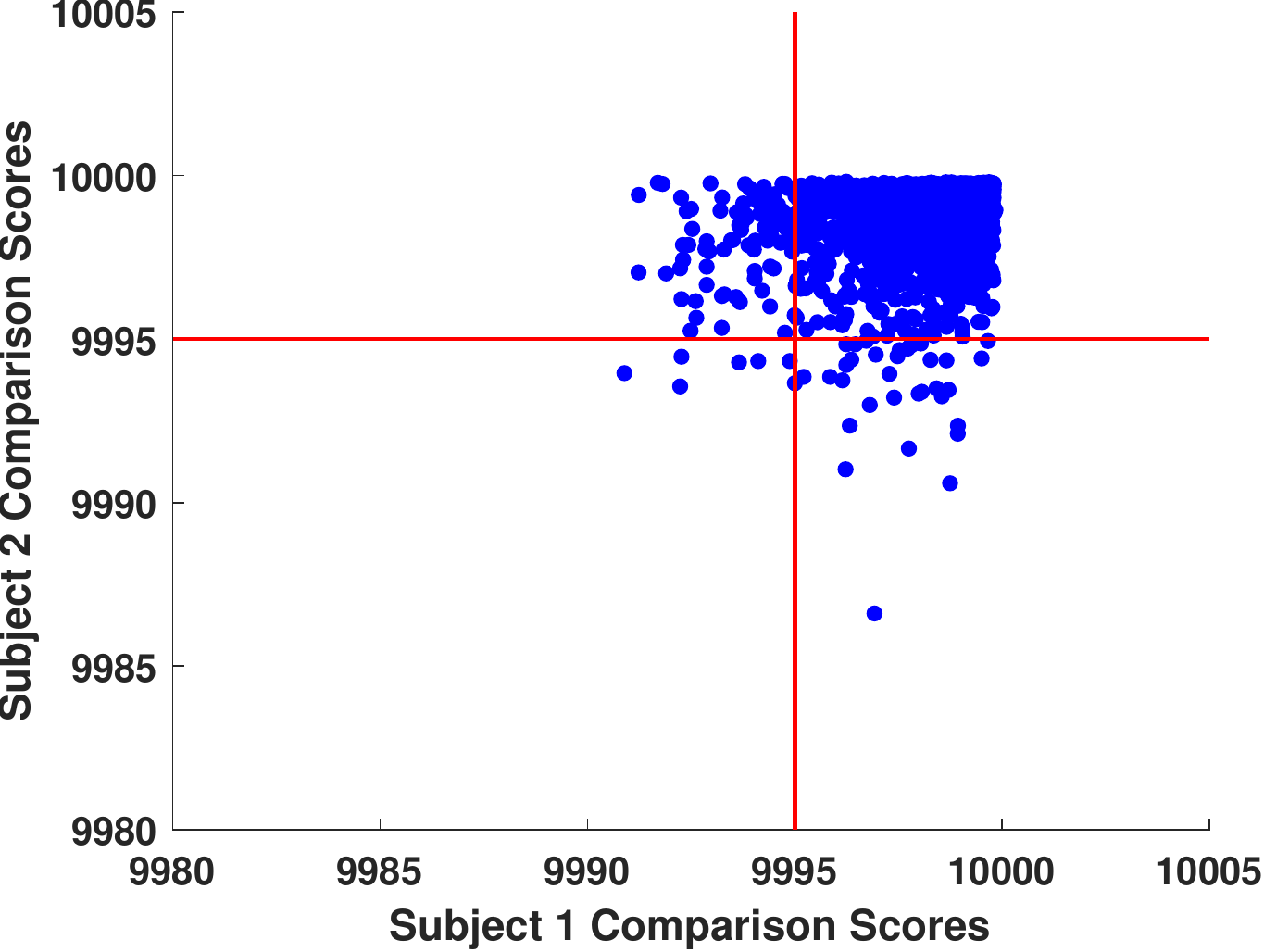}
                \caption{}
                \label{fig:gull8}
        \end{subfigure}%
               \caption{{Vulnerability Plots using 2D \& 3D FRS on Facescape Dataset (a) 2D face FRS using Arcface~\cite{deng2019arcface}, (b) 2D face FRS using COTS, and (c) 3D face FRS  using Led3D~\cite{Led3D2019}, and (d) 3D face FRS  using  Pointnet++~\cite{qi2017pointnet++}}}\label{figRFFiguresProposedFacescape}
\end{figure*}

  \begin{table*}[h!]
    \centering
    \caption{Vulnerability analysis of 2D and 3D FRS on 3D morphing dataset}
      \resizebox{1.0\linewidth}{!}{
    \begin{tabular}{|c|c|c|c|c|c|c|} 
    \hline
    \multicolumn{3}{|c|}{{\bf{Combined}}} & \multicolumn{2}{|c|}{{\bf{Male}}} & \multicolumn{2}{|c|}{{\bf{Female}}}\\ \hline
    Algorithm &  MMPMR\% & FMMPR\% & MMPMR\% & FMMPR\% & MMPMR\% & FMMPR\%\\ \hline
    \multicolumn{7}{|c|}{\bf{2D Vulnerability Analysis}} \\ \hline
    COTS & 97.45\% & 89.78\% & 97.98\% & 90.65\% & 94.03\% & 86.36\%            \\ \hline
    Arcface & 63.81\% & 28.66\%  & 64.92\% & 27.13\% & 59.70\% & 33.33\%   \\ \hline
     \multicolumn{7}{|c|}{\bf{3D Vulnerability Analysis}} \\ \hline
    LED3D~\cite{Led3D2019} & 81.69\% & 54.00\% & 82.67\% & 51.84\% &  77.61\% & 63.64\% \\ \hline
    PointNet++~\cite{qi2017pointnet++} & 95.65\% & 80.52\% & 95.32\% & 79.42\% & 95.52\% & 84.85\%  \\ \hline
    \end{tabular}
    }
    \label{table:vulnerability}
\end{table*}
  
 \begin{table*}[h!]
    \centering
    \caption{{Vulnerability analysis of 2D and 3D FRS on FaceScape Dataset}}
      \resizebox{1.0\linewidth}{!}{
    {\begin{tabular}{|c|c|c|c|c|c|c|} 
    \hline
    \multicolumn{3}{|c|}{{\bf{Combined}}} & \multicolumn{2}{|c|}{{\bf{Male}}} & \multicolumn{2}{|c|}{{\bf{Female}}}\\ \hline
    Algorithm &  MMPMR\% & FMMPR\% & MMPMR\% & FMMPR\% & MMPMR\% & FMMPR\%\\ \hline
    \multicolumn{7}{|c|}{\bf{2D Vulnerability Analysis}} \\ \hline
    COTS & 100\%  & 99.9\% & 100\% & 99.9\% & 100\% &             100\%\\ \hline
    Arcface & 100\%  & 100\%   & 100\% & 100\%  &  100\% & 100\%   \\ \hline
     \multicolumn{7}{|c|}{\bf{3D Vulnerability Analysis}} \\ \hline
    LED3D~\cite{Led3D2019} & 88.8\% & 88.8\% & 90.5\%  & 90.5\%  &  84.9\% & 84.9\% \\ \hline
    PointNet++~\cite{qi2017pointnet++} & 95.4\% & 95.4\%  & 94.1\%  & 94.1\% & 97.5\%  &  97.5\%  \\ \hline
    \end{tabular}
    }}
            \label{table:vulnerability1}
\end{table*}

\subsection{Human Observer Analysis}\label{humanobserver}
We performed human observer analysis to evaluate the human detection performance of the generated 3D morphs.  
The survey is set up online\footnote{\url{https://folk.ntnu.no/jagms}} and is created using PHP, \& HTML-CSS tools. GDPR norms were followed during the survey creation, and only participants' email (used only for registration to avoid duplication), gender, and experience with the morphing problem were recorded. All measures were implemented with full consideration of the participant anonymity. In this study, we designed a GUI for a human observer study to benchmark single-image morphing detection.

Figure \ref{fig:Screenshots} shows a screenshot of the web portal used for the human observer study. The GUI is designed to display two facial images simultaneously, such that one corresponds to the 2D face and the other to the 3D face. Then, the human observer is prompted to independently decide whether these face images are morph or bona fide. Human observers were provided with an option to rotate the 3D face in different directions to make their decisions effectively. Furthermore, opportunities to zoom in and out of the 3D face model are also provided. We have mainly selected to present both 2D/3D face images for human evaluation simultaneously to check whether the 3D information might help detect morphing attacks. Due to the time factor, we used 19 bona fide and 19 morph samples independently from 2D and 3D for the human observer study. Thus, each human observer spent approximately 20 min on average completing this study. Detailed step-wise instructions on using the web portal were available for every participant beforehand.

The human observer study used 36 observers with and without face morphing experience. The quantitative results of the human observer study are shown in Figure \ref{fig:morphingsurvey}. We summarize the human observer's results from the survey as follows.
\begin{itemize}
\item The average detection accuracy of human observers for 2D face bona fide samples is 55.83\% and 42.5\% in a 3D face, respectively. The average detection accuracy of human observers for morphs in 2D  was 58.33\% and 51.85\% for a 3D face. Thus, the detection accuracy is similar for bona fides and morphs in 2D. However, the detection accuracy in 3D was lower for bona fide than for morph.
\item The average detection accuracy is similar for observers without morphing experience and basic morphing experience. Human observers with advanced morphing experience had the highest average detection accuracy. The observers without morphing experience perform similarly to observers with basic morphing experience, which can be attributed to the innate human capacity to distinguish between bona fide  vs. morphed.
\item The survey further validates that generated 3D morphs are challenging to detect from human observations. The average detection accuracy of human observers does not exceed 63.15\%, which shows that the 2D and 3D morphs developed in this study are of high quality and are difficult to detect.
\end{itemize}
The average detection accuracy for a 2D face is higher than that for a 3D face, which can be attributed to the following reasons:
\begin{itemize}
\item The fact that 2D morph is more prevalent, and thus observers generally look for specific artifacts in different regions of the face, makes the task relatively easy with a 2D face.
\item The aspect of what artifacts to look at in 3D is unclear to the human observers, as they are not trained for this task.
\item The quality of generated 3D morphs is high, so human observers find it difficult to distinguish the 3D morphs from the 3D bona fide.
\end{itemize}

\begin{figure*}[htp]
\begin{center}
\includegraphics[width=0.9\linewidth]{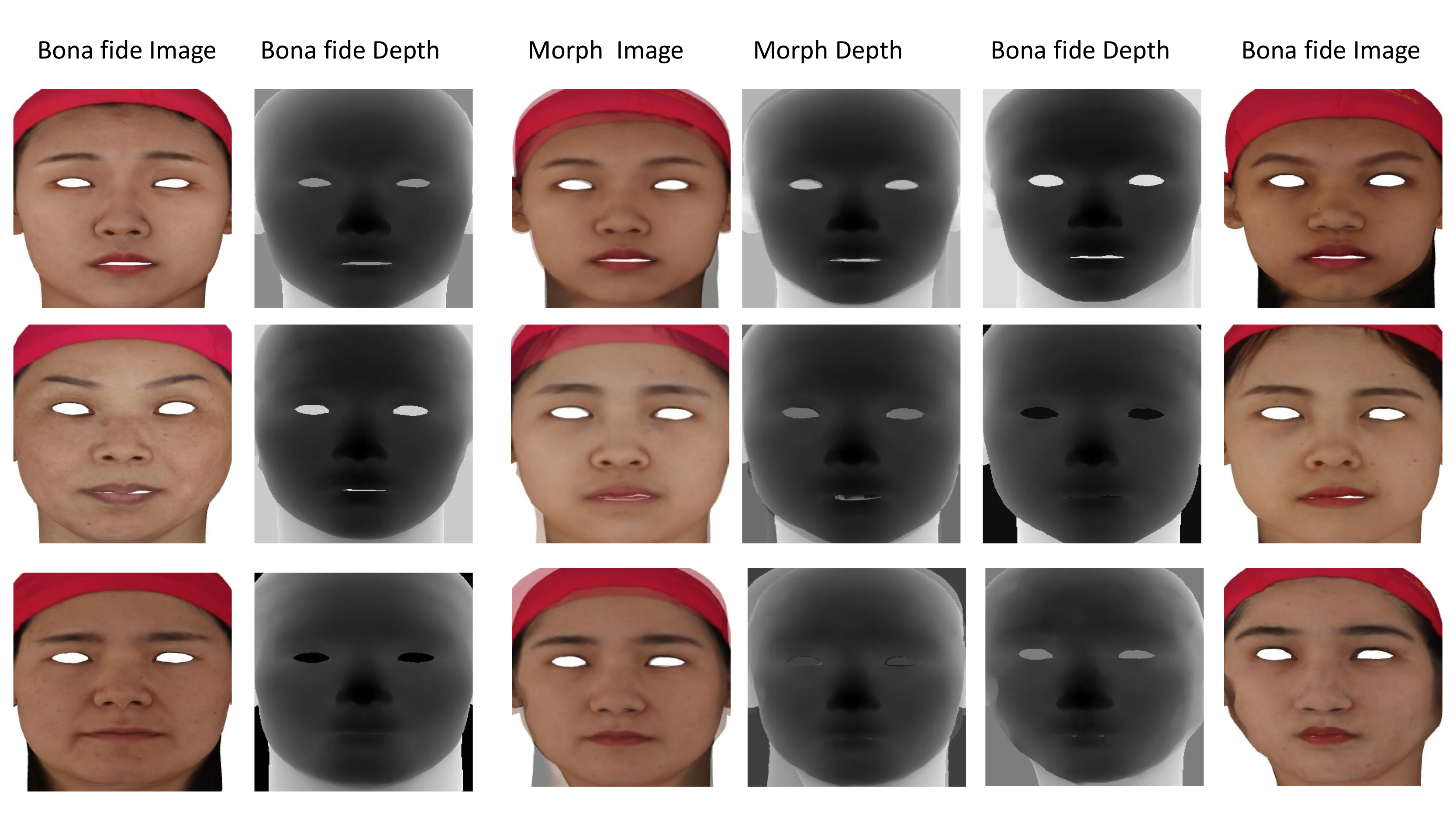}
\end{center}
   \caption{ { {{Illustration of the Color Images and Depth Maps of Bona fide Samples and Face Morphs generated using the proposed method on Facescape Dataset ~\cite{Yang_2020_Facescape}}}}}
\label{fig:sotaComparisonQualitativeResponse1}
\end{figure*}
\subsection{Vulnerability Study}
\label{sec:vul}
In this work, we benchmarked the performance of  automatic FRS on both 2D and 3D face models. The 2D face vulnerability was computed using the color image, and the 3D face vulnerability was calculated based on the depth map/point cloud. We have used two different metrics to benchmark the vulnerability assessment : the  Mated Morphed Presentation Match Rate (MMPMR)~\cite{MMPMRUlrich} and Fully Mated Morphed Presentation Match Rate (FMMPMR)~\cite{FMMPMRSushma}. { {MMPMR can be defined as the percentage of morph samples, which can be verified with all contributing data subjects~\cite{FMMPMRSushma}. However, MMPMR does not consider the number of attempts made during the score computation. This is rectified in FMMPMR~\cite{FMMPMRSushma}, where the morphed image sample should be verified across all attempts. The higher value of MMPMR and FMMPMR indicates the higher vulnerability of the FRS.}} Vulnerability analysis was performed by enrolling the morphing image (2D/3D) and then obtaining the comparison score by probing both contributory data subjects' facial images (2D/3D). To compute the vulnerability of 2D face morphing images, we used two different FRS:  Arcface~\cite{deng2019arcface} and a commercial-off-the-shelf (COTS) FRS \footnote{The name of the COTS is not indicated to respect confidentiality}. 3D face vulnerability analysis uses Deep Learning-based FRS such as Led3D~\cite{Led3D2019} and PointNet++~\cite{qi2017pointnet++}. The thresholds for all FRS used in this study were set at FAR=0.1\%, following the guidelines of Frontex for border control \cite{frontex}.  

\subsubsection{Quantitative vulnerability results on 3D morphing dataset}
The results are summarized in Table~\ref{table:vulnerability} and the vulnerability plots are presented in Figure~\ref{figRFFiguresProposed}. Based on the obtained results, it can be noted that (1) both 2D and 3D FRS are vulnerable to the generated face morphing attacks, and (2) among the 2D FRS, COTS indicates the highest vulnerability compared to Arcface FRS. (3) Among the 3D FRS, PointNet++~\cite{qi2017pointnet++} indicates the highest vulnerability. Thus, the quantitative results of the vulnerability analysis indicate the effectiveness of the generated 3D face morphing attacks. 

\subsubsection{Quantitative vulnerability results on Facescape dataset}
We used 100 unique databases with 56 male and 44 female subjects. For each subject, we selected two 3D face scans. One was used to generate the 3D face morphing and the other was used as the probe image to obtain the comparison score to compute the vulnerability metrics. The proposed method was then used to obtain 3D morphing models, resulting in 2486 morphing models. Figure \ref{fig:sotaComparisonQualitativeResponse1} shows an example of the proposed 3D morphing generation samples together with bona fide 3D scans from the Facescape Dataset ~\cite{Yang_2020_Facescape}. The quantitative vulnerability results for the escape dataset are listed in Table  \ref{table:vulnerability1}, and the vulnerability plots are shown in Figure~\ref{figRFFiguresProposedFacescape}. In addition, it can be observed that the proposed 3D face morphing generation samples exhibit high vulnerability with both 2D and 3D FRS. Among the 2D FRS, both COTS and Arcface indicate similar vulnerabilities with MMPMR = 100\%. However, among the 3D FRS, PointNet++~\cite{qi2017pointnet++} shows the highest vulnerability.

Thus, based on the vulnerability analysis reported on the 3DMD and Facescape datasets with 2D and 3D FRS, the proposed 3D face morphing technique indicates consistently high vulnerability. The vulnerability is noted to be high with the Facescape dataset compared with the 3D morphing dataset. The variation in the vulnerability performance across different FRS  can be attributed to the type of feature extraction and classification techniques employed in individual FRS. For example, 2D face recognition systems are based on identity features, whereas 3D-based systems are based on high-resolution depth and shape.

\subsection{Automatic 3D Face Point Cloud Quality Estimation}
\label{sec:qulaity}

In this work, we estimate the visual quality based on the effectiveness of different types of features, including both color and geometry, as proposed in  \cite{zhang2021noreference}. This study aimed to quantitatively estimate the quality of the generated 3D face morphing point clouds and bona fide 3D face point clouds to quantify the quality of the proposed morphing generation. To this extent, five different point cloud features based on geometry, namely curvature, anisotropy, linearity, planarity, and sphericity, and three color information features, namely L color component, A color component, and B color component, were computed to benchmark the quality based on the geometry of the generated 3D morphing models. 

\begin{figure*}[th!]
\centering
 \begin{tabular}[b]{c}
       \includegraphics[width=0.18\linewidth]{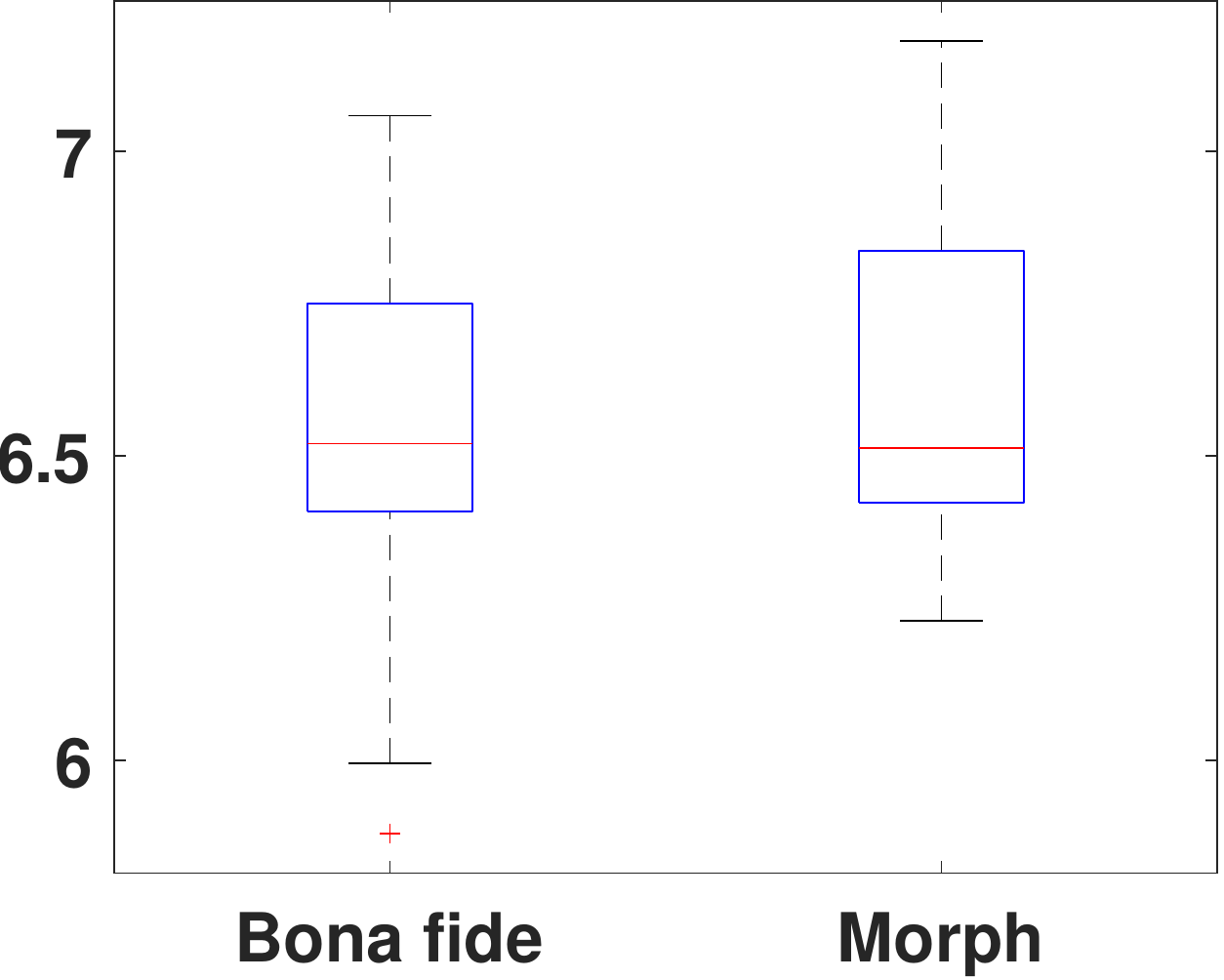}\\ 
    \small {(a) L Color Component}
  \end{tabular}
  \begin{tabular}[b]{c}
       \includegraphics[width=0.18\linewidth]{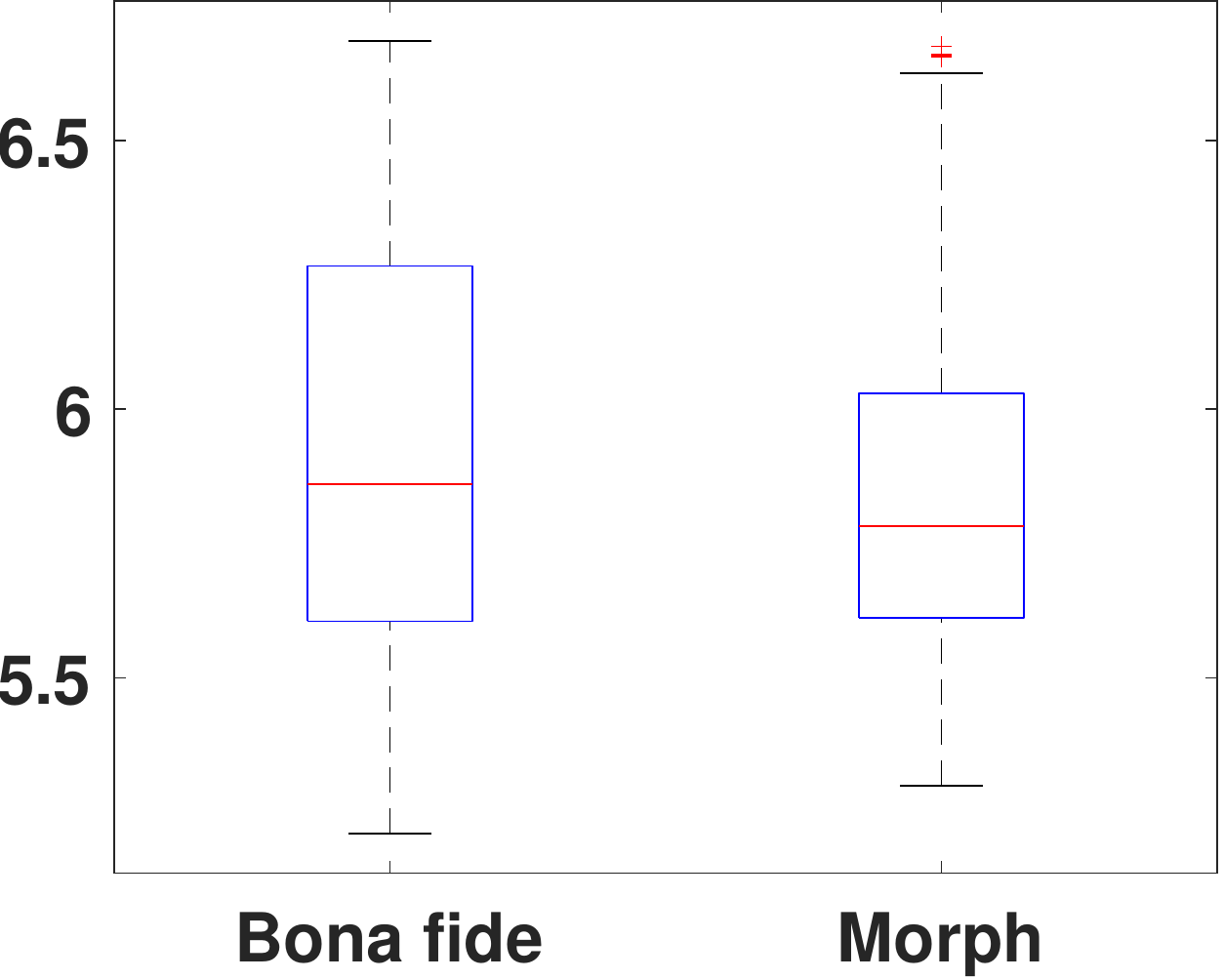}\\ 
    \small {(b) A Color Component}
  \end{tabular}
\begin{tabular}[b]{c}
       \includegraphics[width=0.18\linewidth]{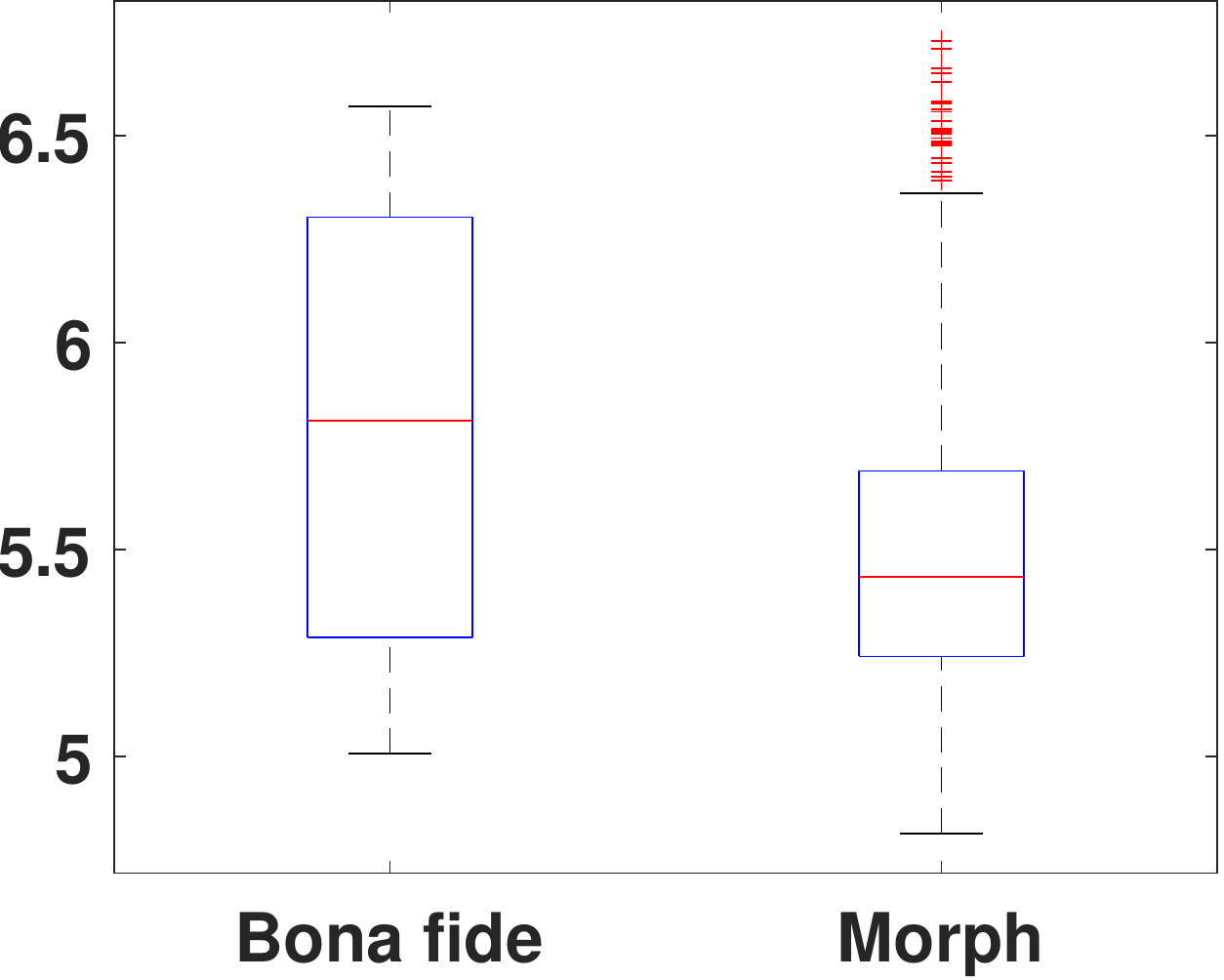}\\ 
    \small {(c) B Color Component}
  \end{tabular}
\begin{tabular}[b]{c}
       \includegraphics[width=0.18\linewidth]{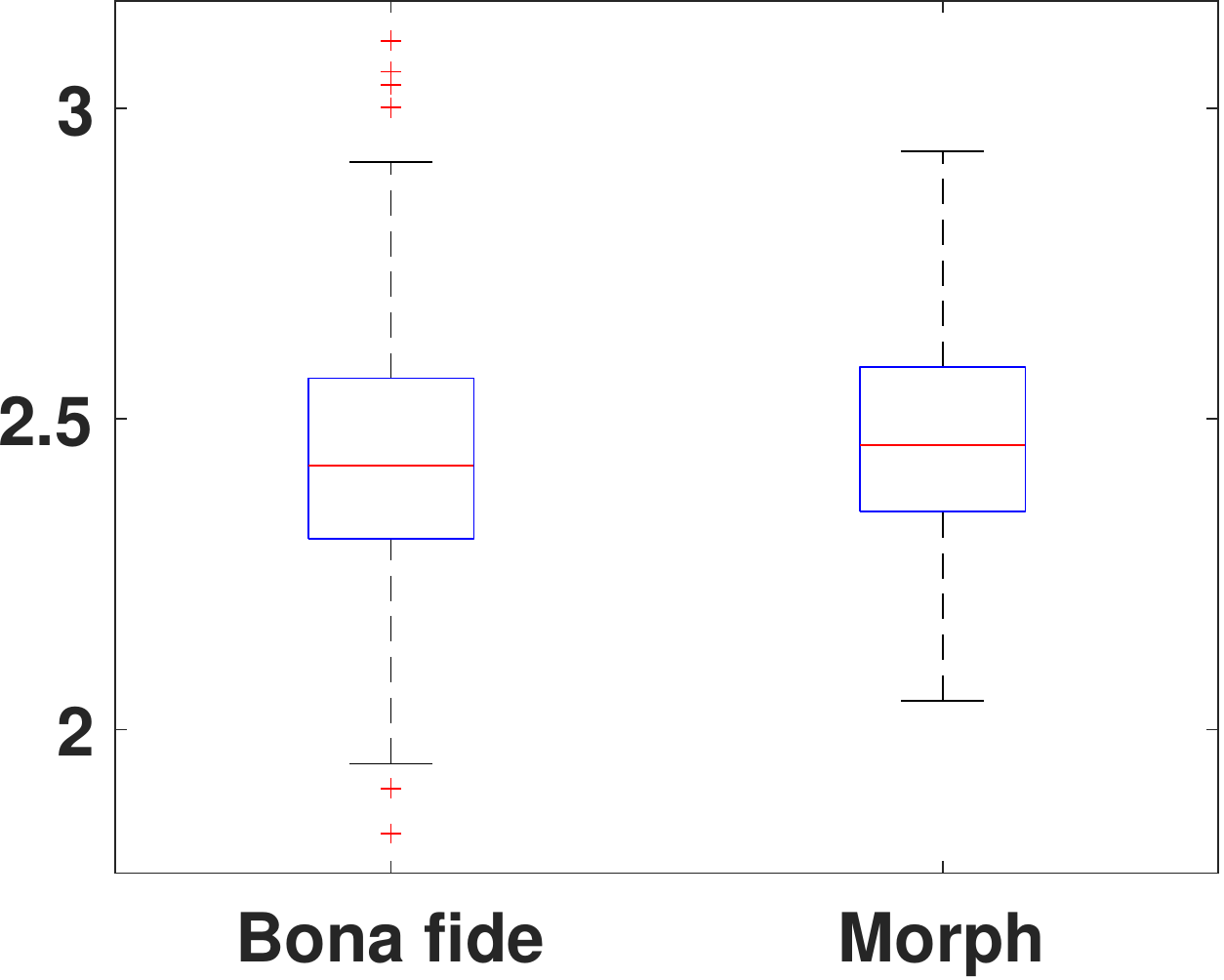}\\
    \small {(d) Planarity}
  \end{tabular}
\begin{tabular}[b]{c}
       \includegraphics[width=0.18\linewidth]{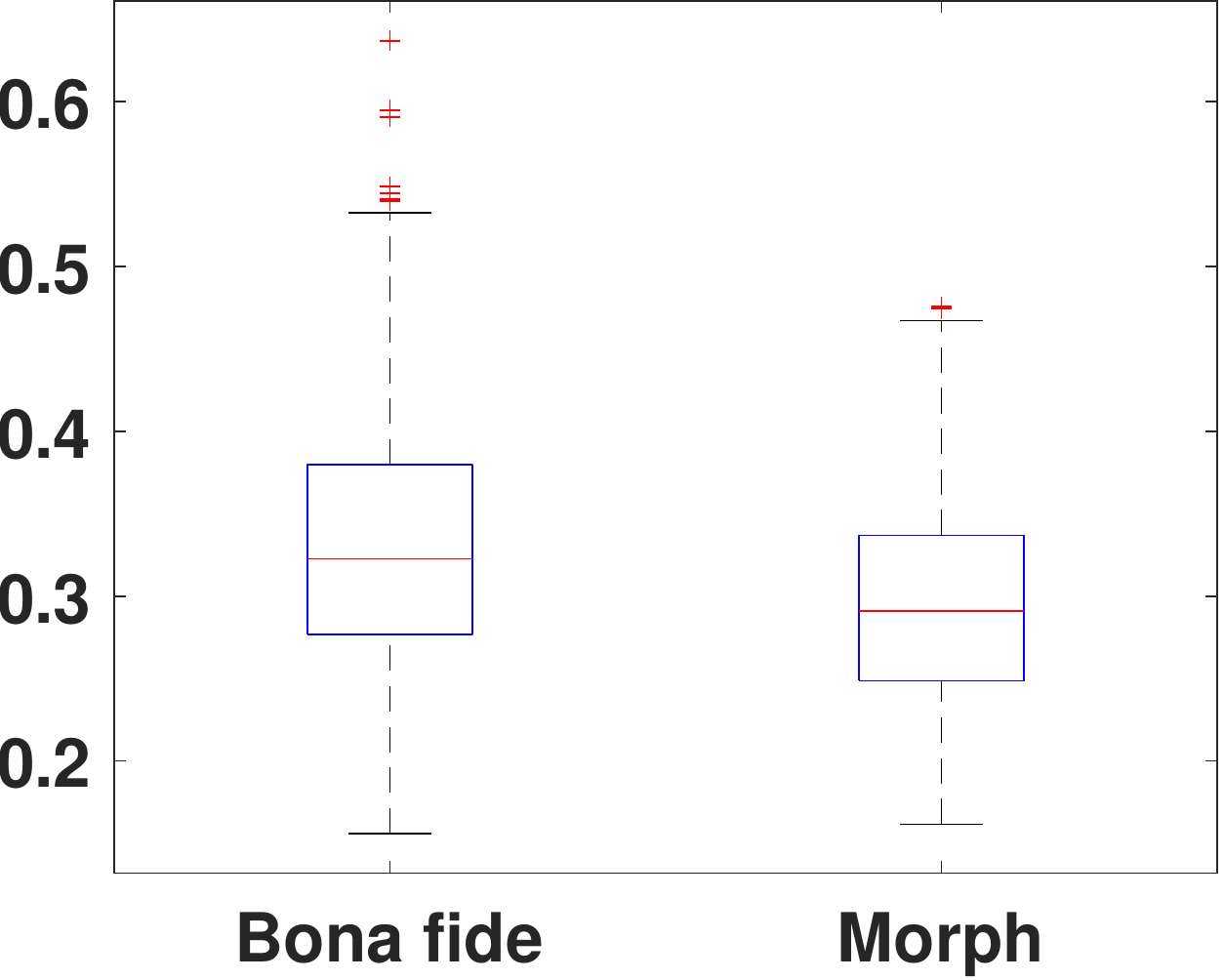}\\
    \small {(e) Sphericity}
  \end{tabular}
  \begin{tabular}[b]{c}
       \includegraphics[width=0.18\linewidth]{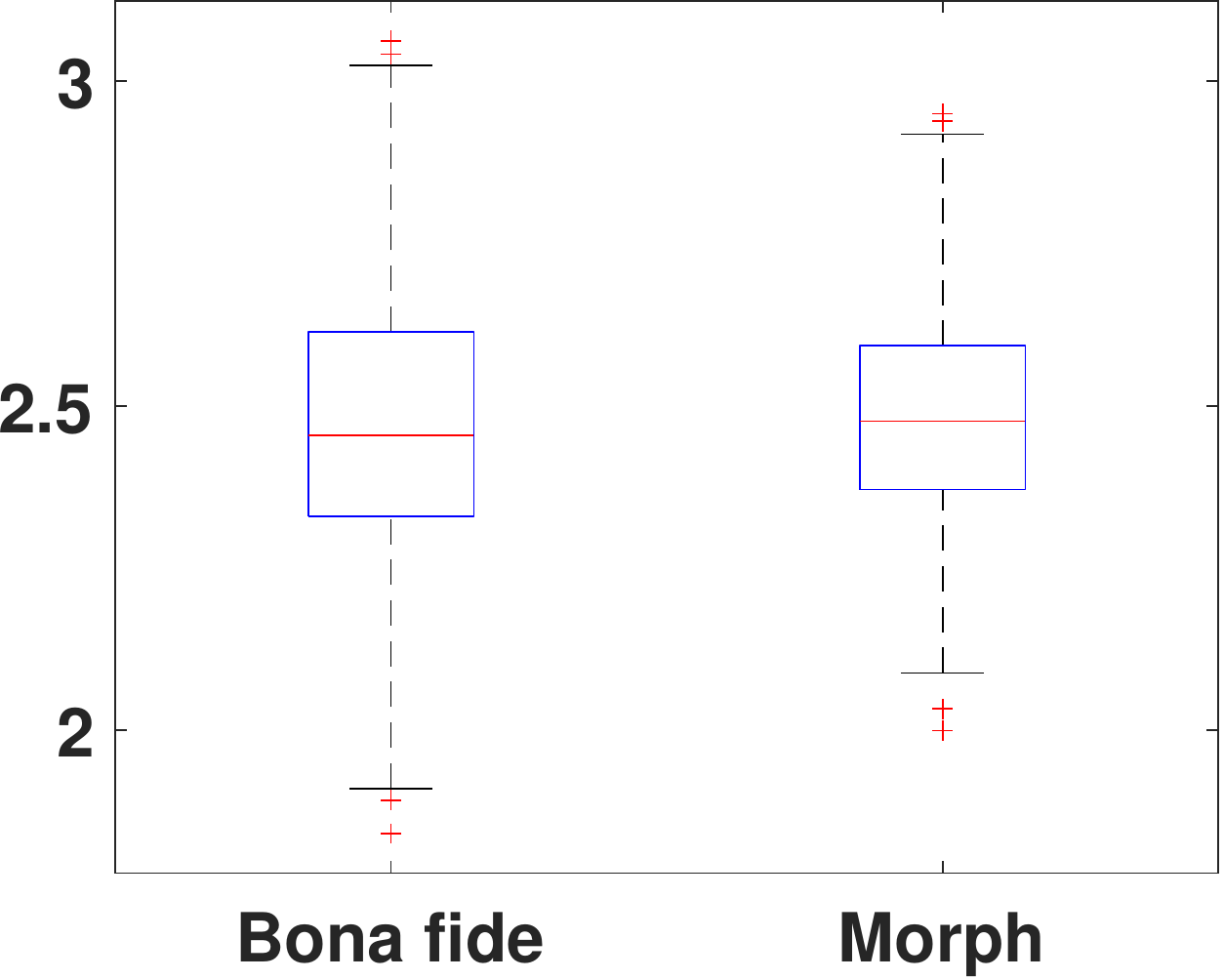}\\
    \small {(f) Linearity}
  \end{tabular}
   \begin{tabular}[b]{c}
       \includegraphics[width=0.18\linewidth]{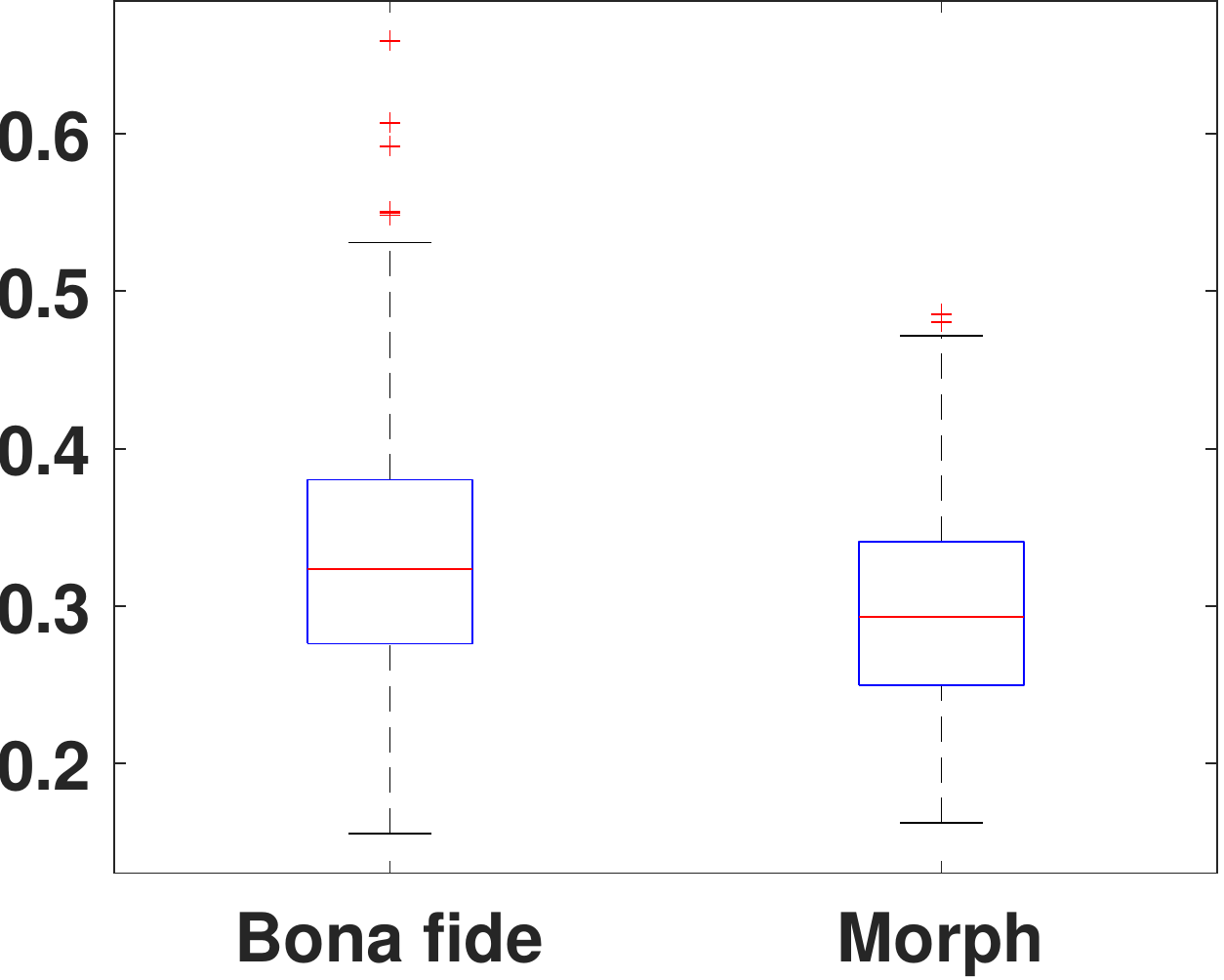}\\
    \small {(g) Curvature}
  \end{tabular}
  \begin{tabular}[b]{c}
       \includegraphics[width=0.18\linewidth]{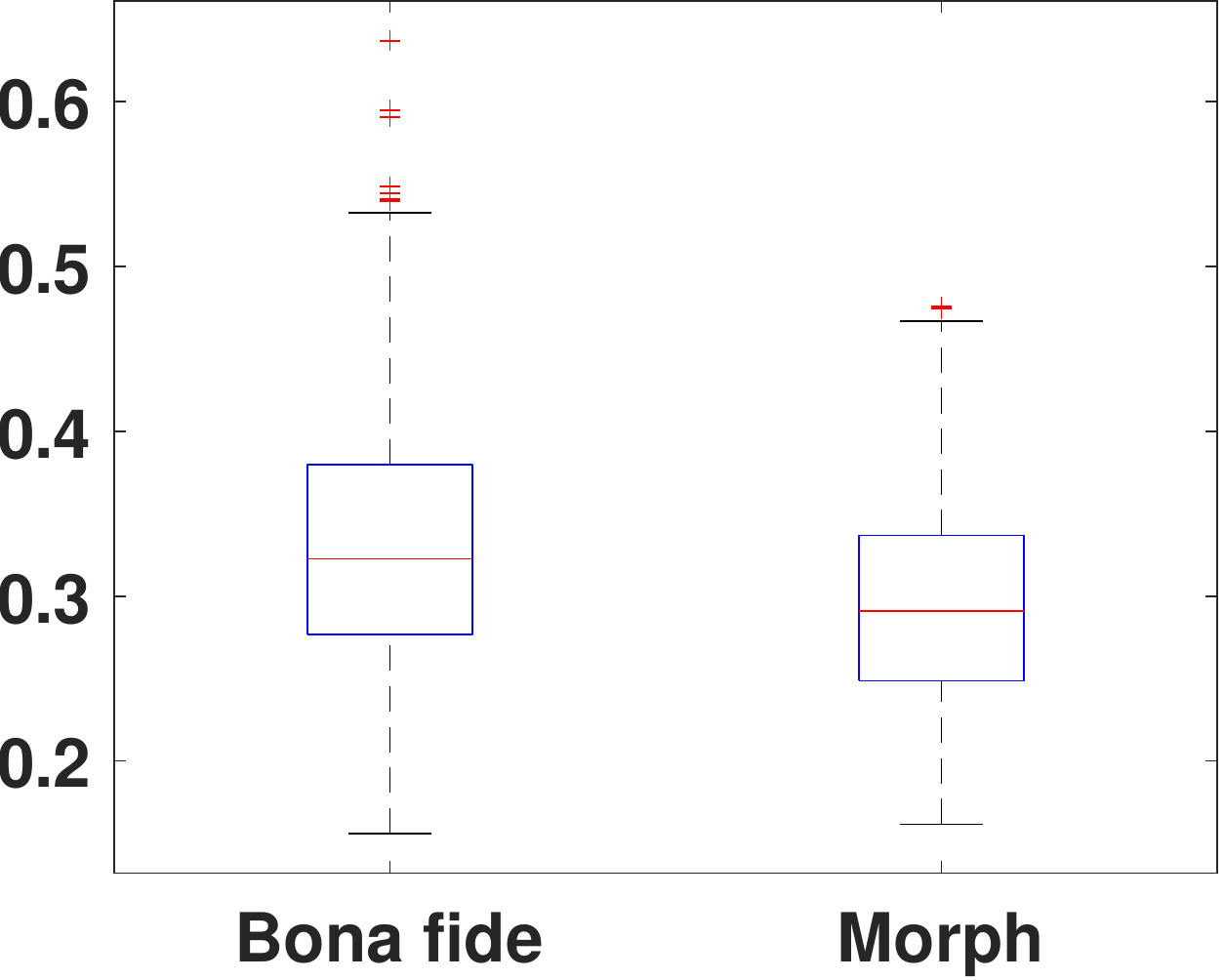}\\
    \small {(h) Anisotropy}
  \end{tabular}
     \caption{Box plots showing the eight different 3D model quality estimation from 3D bona fide and 3D morph based on color and geometry}
     \label{figRFFiguresPointCloudQuality}
\end{figure*}

\begin{table}[htp]
    \centering
    \caption{Quantitative values of quality features for 3D face point clouds corresponding to 3D bona fide and morph based on color and geometry}
    \resizebox{1\linewidth}{!}{
    \begin{tabular}{|c|c|c|}
    \hline
    \multicolumn{1}{|c|}{{\bf{3D Face Quality Features (mean $\pm$ std. deviation)}}} & 
    \multicolumn{2}{|c|}{{\bf{Data type}}}    \\ \hline
    & \multicolumn{1}{|c|}{{\bf{Bona fide}}} & 
    \multicolumn{1}{|c|}{{\bf{Morphed}}} \\ \hline
    {\bf{L Color}} & 6.5614$\pm$0.2191  & 6.6076$\pm$0.2340 \\ \hline
    {\bf{A Color}} & 5.9368$\pm$0.3547 & 5.8546$\pm$0.3260 \\ \hline
    {\bf{B Color}} & 5.7998$\pm$0.5074 &  5.5326$\pm$0.4198 \\ \hline
    {\bf{Linearity}} & 2.4708$\pm$0.2196 & 2.4911$\pm$0.1776 \\ \hline
    {\bf{Sphericity}} & 0.3318$\pm$0.0807 & 0.2936$\pm$0.0592 \\ \hline
    {\bf{Anisotropy}} & 0.3318$\pm$0.0807 &  0.2936$\pm$0.0592 \\ \hline
    {\bf{Curvature}} & 0.3330$\pm$0.0821 & 0.2965$\pm$0.0606 \\ \hline
    {\bf{Planarity}} &  2.4430$\pm$0.2176 & 2.4711$\pm$0.1733 \\ \hline
 \end{tabular}
 }
             
             \label{table:qualitydetails}

\end{table} 
Figure \ref{figRFFiguresPointCloudQuality} shows the box plot of the eight different quality metrics for both 3D bona fide and 3D morphing point clouds. The quantitative values (mean and standard deviation) of different quality features are also shown in Table \ref{table:qualitydetails}. As noted from Figure \ref{figRFFiguresProposed}, the quality estimations, mainly based on geometry, indicate the near-complete overlapping for 3D bona fide and 3D morph. Thus, the proposed 3D face morphing generation did not degrade the depth quality. Instead, it has achieved comparable quality based on geometry from bona fide 3D models used for the morphing operation. A similar observation can also be noted with the color image quality estimation. 

\subsection{3D Face Morphing Attack Detection}\label{3DFMAD}
\begin{figure}[htp]
\begin{center}
\includegraphics[width=0.9\linewidth]{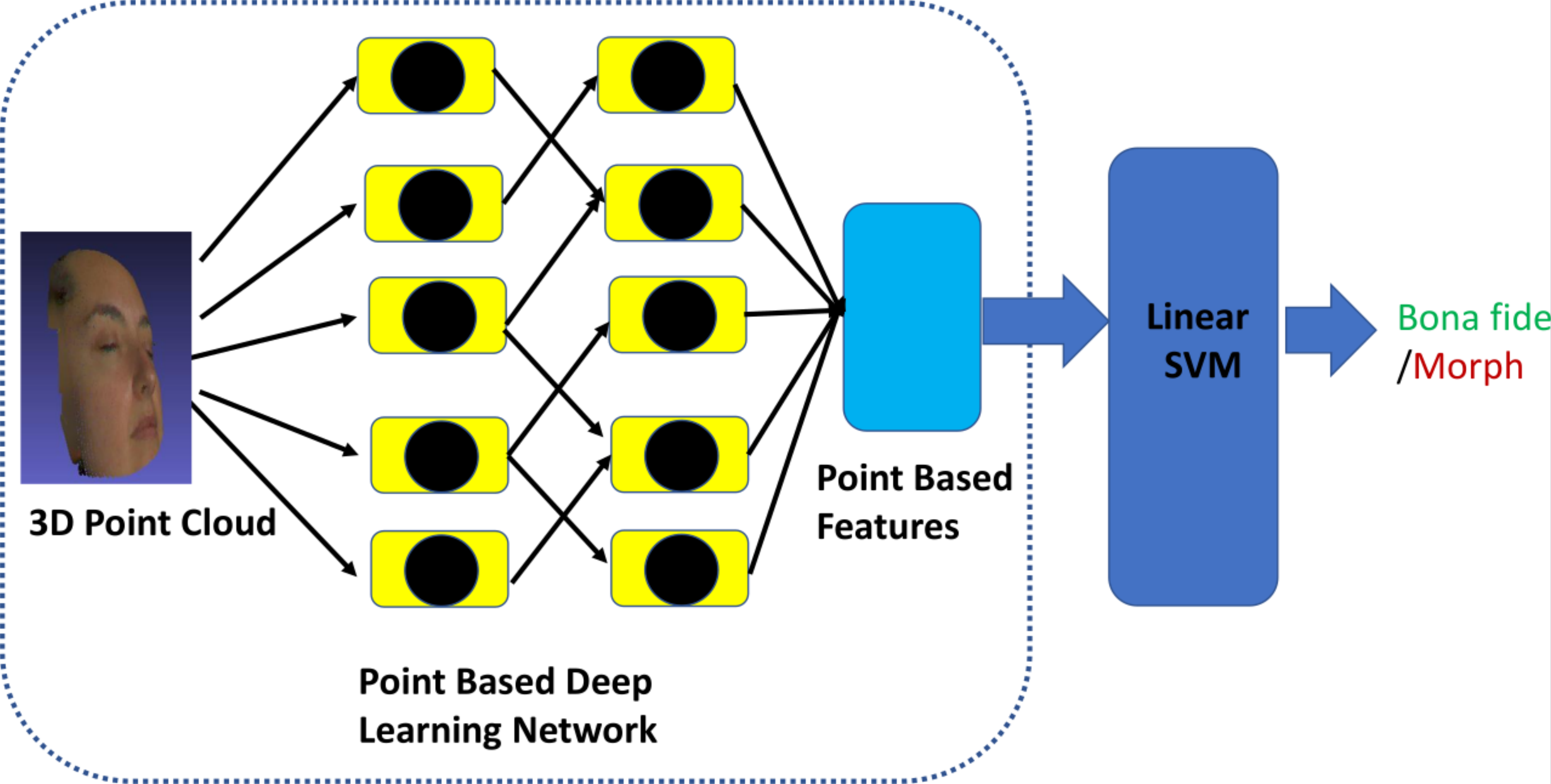}
\end{center}
   \caption{Illustration of the proposed 3D face MAD}
\label{fig:maddetection}
\end{figure}

In this section, we present our proposed method for a single 3D model-based MAD. Because 3D face morphing is extensively presented in this paper for the first time, there is no state-of-the-art method for detecting these attacks. Therefore, we were motivated to develop 3D MAD techniques to reliably detect these attacks. 
The proposed 3D MAD techniques are based on pre-trained 3D point-based networks used to extract features, as shown in Figure \ref{fig:maddetection}. Thus, given the 3D face point clouds, we first computed the features from the pre-trained network, and in the next step, we fed the same to the linear support vector machine to make the final decision on either bona fide or morph. In this work, we used three different pre-trained point networks, namely   Pointnet~\cite{qi2017pointnet,goyal2021revisiting}, Pointnet++~\cite{qi2017pointnet++,goyal2021revisiting} and SimpleView~\cite{goyal2021revisiting}, to benchmark the 3D MAD performance. All three pretrained CNNs were trained on the ModelNet40 dataset~\cite{Modelnet40}.

Pointnet~\cite{qi2017pointnet,goyal2021revisiting} was one of the earliest point-based classifications of deep learning networks invariant to the permutation of 3D vertices. Given the 3D face point clouds,  we extracted the features from the classification task layer corresponding to a feature dimension of 4096. Pointnet++~\cite{qi2017pointnet++,goyal2021revisiting} is an improved version of Pointnet~\cite{qi2017pointnet,goyal2021revisiting} that was achieved by introducing a hierarchical neural network that was applied recursively. In this work, given the 3D face point clouds, we extracted the features from the classification task layer of Pointnet++  to obtain a 40-dimensional feature vector. SimpleView~\cite{goyal2021revisiting} network is based on projecting point clouds onto multiple-view depth maps. In this work, given the 3D face point clouds, we extract the features from the classification task layer of the SimpleView network to obtain a 40-dimensional feature vector. 

\begin{figure}[htp]
\begin{center}
\includegraphics[width=0.75\linewidth]{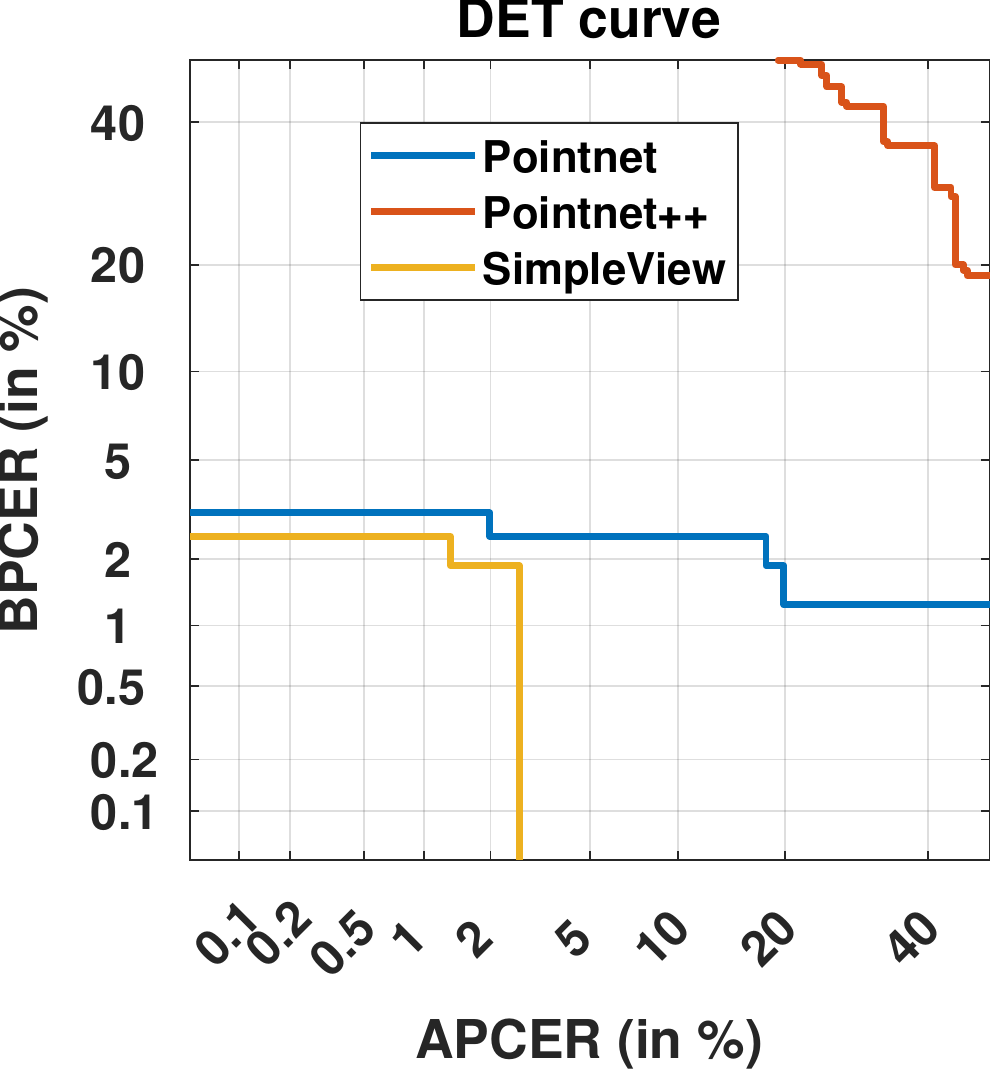}
\end{center}
   \caption{DET Curve for the Proposed 3D Morphing Detection methods.}
\label{fig:morphingdetection}
\end{figure}

To effectively benchmark the performance of the proposed 3D MAD, we divided the newly collected dataset into two independent sets: training and testing. The training set consisted of 3D bona fide and morphing samples from 21 unique data subjects, whereas the testing set consisted of 3D samples from 20 unique data subjects. Thus, the training set consisted of 168 bona fide and 194 morphed features and the testing set consisted of 160 bona fide and 151 morphed features, as summarized in Table \ref{table:detectiondataset}. Table \ref{table:detection} presents the quantitative performance of the proposed 3D MAD technique. Figure \ref{fig:morphingdetection} shows the performance of the individual algorithms in DET. The performance is benchmarked using ISO/IEC metrics \cite{ISO2017} defined as the Attack Presentation Classification Error Rate (APCER), which is the misclassification rate of attack presentations, and the bona fide Presentation Classification Error Rate (BPCER), which is the misclassification of bona fide presentation as attacks. Based on the results, the best performance is obtained with the SimpleView~\cite{goyal2021revisiting} network with a D-EER of 1.59\%.

\begin{table}[h!]
    \centering
     \caption{Morphing Attack Detection (S-MAD) Method Protocol}
    \resizebox{0.8\linewidth}{!}
    {
    \begin{tabular}{|c|c|c|c|} 
\hline
\multicolumn{4}{|c|}{ Train Dataset (21 Subjects)} \\ \hline
\multicolumn{2}{|c|}{Bona fide Samples} & \multicolumn{2}{|c|}{Morphing Samples}   \\ \hline
\multicolumn{2}{|c|}{168}  & \multicolumn{2}{|c|}{194} \\ \hline
\multicolumn{4}{|c|}{ Test Dataset (20 Subjects)} \\ \hline
\multicolumn{2}{|c|}{Bona fide Samples} & \multicolumn{2}{|c|}{Morphing Samples}   \\ \hline
\multicolumn{2}{|c|}{160}  & \multicolumn{2}{|c|}{151} \\ \hline

    \end{tabular}
}\label{table:detectiondataset}
    \end{table}
\begin{table}[h!]
    \centering
     \caption{Quantitative performance of the proposed 3D MAD techniques}
    \resizebox{1.0\linewidth}{!}
    {
    \begin{tabular}{|c|c|c|c|} 
\hline
    Algorithm & {D-EER (\%)}  & \multicolumn{2}{|c|}{BPCER @ APCER =} \\ \cline{3-4}
Proposed Method & & {5\%} & {10\%} \\ \hline \hline
    Pointnet~\cite{qi2017pointnet} & {2.57} & {3.12} & {2.5} \\ \hline
    Pointnet++~\cite{qi2017pointnet++} & {37.33} & {81.87} & {68.12} \\ \hline
    SimpleView~\cite{goyal2021revisiting} & {\bf{1.59}} & {\bf{2.5}} & {\bf{0}} \\ \hline
    \end{tabular}
    }
   
    \label{table:detection}
    \end{table}
\section{Discussion}
\label{sec:discussion}
Based on the extensive experiments and obtained results made above, the research questions formulated in Section \ref{intro} are answered below.
\begin{itemize}[leftmargin=*,noitemsep, topsep=0pt,parsep=0pt,partopsep=0pt]
	\item {\textbf{RQ\#1}. Does the proposed 3D face morphing generation technique yield a high-quality 3D morphed model?}
	\begin{itemize}[leftmargin=*,noitemsep, topsep=0pt,parsep=0pt,partopsep=0pt]
		\item Yes, the proposed method of generating the 3D face morphing has resulted in a high-quality morphed model almost similar to that of the original 3D bona fide. The quality analysis reported in Figure \ref{figRFFiguresPointCloudQuality} and Table \ref{table:qualitydetails} also justifies the quality of the generated 3D morphs quantitatively as the quality values from 3D morphing show larger overlapping with the 3D bona fide. In addition, the human observer analysis reported in Section \ref{humanobserver} also justifies the quality of the proposed 3D face morphing generation method as it is found reasonably difficult to detect based on the artefacts. 
	\end{itemize}
	\item {\textbf{RQ\#2}. Does the generated 3D face morphing model indicate the vulnerability for both automatic 3D FRS and human observers? } 
	\begin{itemize}[leftmargin=*,noitemsep, topsep=0pt,parsep=0pt,partopsep=0pt]
		\item  Yes, based on the analysis reported in Section \ref{sec:vul}, the generated 3D face morphing model indicates a high degree of vulnerability for both automatic 3D FRS and human observers.
	\end{itemize}
	\item {\textbf{RQ\#3}. Are the generated 3D face morphing models more vulnerable when compared to 2D face images for both automatic 3D FRS and human observers?}
	\begin{itemize}[leftmargin=*,noitemsep, topsep=0pt,parsep=0pt,partopsep=0pt]
		\item Equally vulnerable, the 3D face morphing models are more vulnerable than their 2D counterparts, as shown in Figure~\ref{figRFFiguresProposed} when using automatic FRS. 
		\item However, the vulnerability is almost comparable when evaluated by a human observer study (see Section~\ref{humanobserver}), where one of the main reasons could be more prevalence of 2D morphs, which makes human observers sensitive about which artifacts to look for.
	\end{itemize} 
	
	\item {\textbf{RQ\#4}.Can the 3D point cloud information be used to detect the 3D face morphing attacks reliably?}
	\begin{itemize}[leftmargin=*,noitemsep, topsep=0pt,parsep=0pt,partopsep=0pt]
		\item Yes, on using the proposed 3D face morphing attack Detection approaches (see Section~\ref{3DFMAD}) the point cloud information can be used for reliable 3D morphing detection. 
		
	\end{itemize}  
	
\end{itemize}

\vspace{-4mm}
\section{Limitations of Current Work and Potential Future Works}
\label{sec:limitations-future-works}
Although this work presents a new dimension for face morphing attack generation and detection, especially in 3D, it has a few limitations. In the current scope of work, 3D morph generation and detection were carried out on high-quality 3D scans collected using the Artec Eva sensor. We employed high-quality 3D face scans to achieve good enrolment quality scans that may reflect real-life ID enrolment scenarios. Thus, future studies should investigate the proposed 3D morphing generation and detection techniques using low-quality (depth) 3D scans. Furthermore, extending the study to in-the-wild capture can also be considered in future work. Second, the analysis was conducted using 41 data subjects due to the present pandemic outbreak. However, we also present the results on the publicly available 3D face dataset, Facescape, with 100 unique IDs. 
Future work could benchmark the proposed method on large-scale datasets with different 3D resolutions. Third, cleaning the noise from 3D scans is tedious and sometimes requires manual intervention. Thus, future work can develop fully automated noise removal methods in 3D point clouds to easily generate 3D morphs.

\vspace{-4mm}
\section{Conclusion}\label{conclusion}

This work presented a new dimension for face morphing attack generation and detection, particularly in 3D. We introduced a novel algorithm to generate high-quality 3D face morphing models using point clouds. To validate the attack potential of the newly generated 3D face-morphing attacks, vulnerability analysis uses 2D and 3D FRS. Furthermore, human observer analysis is presented to investigate the usefulness of 3D information in morph detection. The obtained results justify the high vulnerability of the proposed 3D face morphing models. We also presented an automatic quality analysis of the generated 3D morphing models, which indicated a similar quality to the bona fide 3D scans. Finally, we proposed three different 3D MAD algorithms to detect 3D morphing attacks using pretrained point-based CNN models. Extensive experiments indicated the efficacy of the proposed 3D MAD algorithms in detecting 3D face-morphing attacks.

{\vspace{-2mm}
\small
\balance
\bibliographystyle{IEEEtran}
\bibliography{main}
}

\begin{IEEEbiography}[{\includegraphics[width=1in,height=1.25in,clip,keepaspectratio]{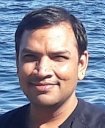}}]{Jag Mohan Singh} (Member, IEEE) received the
B.Tech. (Hons.) and M.S. by research in computer
science degrees from the International Institute
of Information Technology (IIIT), Hyderabad, in
2005 and 2008, respectively. He is currently in the
final year of his Ph.D. with the Norwegian Biometrics
Laboratory (NBL), Norwegian University
of Science and Technology (NTNU), Gjøvik. He
worked with the industrial research and development
departments of Intel, Samsung, Qualcomm,
and Applied Materials, India, from 2010 to 2018. He has published several
papers at international conferences focusing on presentation attack detection,
morphing attack detection and ray-tracing. His current research interests
include generalizing classifiers in the cross-dataset scenario and neural
rendering.
\end{IEEEbiography}

\begin{IEEEbiography}[{\includegraphics[width=1in,height=1.25in,clip,keepaspectratio]{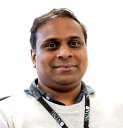}}]{Raghavendra Ramachandra} 
(Senior Member, IEEE) obtained a
Ph.D. in computer science and technology from
the University of Mysore, Mysore India and Institute
Telecom, and Telecom Sudparis, Evry, France
(conducted as collaborative work) in 2010. He
is currently a full professor at the Institute of
Information Security and Communication Technology
(IIK), Norwegian University of Science
and Technology (NTNU), Gjøvik, Norway. He is
also working as R\&D chief at MOBAI AS. He
was a researcher with the Istituto Italiano di Tecnologia, Genoa, Italy,
where he worked on video surveillance and social signal processing.
His main research interests include deep learning, machine learning, data
fusion schemes, and image/video processing, with applications to biometrics,
multi-modal biometric fusion, human behaviour analysis, and crowd
behaviour analysis. He has authored several papers and is a reviewer for
Several international conferences and journals. He also holds several patents
for the biometric presentation attack detection and morphing attack detection, respectively.
He has also been involved in various conference organising and program
committees, and has served as an associate editor for various journals. He
has participated (as a PI, co-PI or contributor) in several EU projects, IARPA
USA, and other national projects. He is serving as an editor of the ISO/IEC
24722 standards on multi-modal biometrics and an active contributor to the
ISO/IEC SC 37 standards for biometrics. He has received several best paper
awards. He is also a senior member of the IEEE and VP of Finance at the IEEE Biometric Council.

\end{IEEEbiography}

\end{document}